\pdfoutput=1

\documentclass[nohyperref]{article}
\PassOptionsToPackage{hyphens}{url}

\usepackage{microtype}
\usepackage{graphicx}
\usepackage{booktabs} 

\usepackage{hyperref}

\usepackage[accepted]{icml2022}

\usepackage{amsmath}
\usepackage{amssymb}
\usepackage{mathtools}
\usepackage{amsthm}
\usepackage{upquote}
\usepackage{listings}
\usepackage{color}
\usepackage{caption}

\usepackage[capitalize,noabbrev]{cleveref}

\usepackage{wrapfig}
\usepackage{float}
\usepackage{subcaption}
\usepackage{adjustbox}
\usepackage{enumitem}
\setitemize{itemsep=1pt, topsep=1pt, parsep=4pt}
\usepackage{multirow}
\usepackage{tikz}
\usetikzlibrary{positioning}
\usetikzlibrary{arrows}
\usetikzlibrary{arrows.meta}
\usetikzlibrary{patterns}
\usepackage{pgfplots}
\usepackage{pgfplotstable}
\usepgfplotslibrary{groupplots}

\newcommand{\ie}{i.e.}
\newcommand{\eg}{e.g.}

\newcommand{\resp}{resp.}

\newcommand{\vs}{v.s.}
\newcommand{\code}[1]{\texttt{\small{#1}}}
\newcommand{\mytexttt}[1]{\texttt{\small #1}}

\newcommand{\varmisuse}{\mytexttt{var-misuse}}
\newcommand{\wrongbinop}{\mytexttt{wrong-binop}}
\newcommand{\argswap}{\mytexttt{arg-swap}}
\newcommand{\nobug}{\mytexttt{NO\_BUG}}
\newcommand{\cubert}{CuBERT}
\newcommand{\gnn}{GNN}
\newcommand{\great}{GREAT}
\newcommand{\buglab}{BugLab}
\newcommand{\strain}{\mytexttt{syn-train}}
\newcommand{\rtrain}{\mytexttt{real-train}}
\newcommand{\rval}{\mytexttt{real-val}}
\newcommand{\rtest}{\mytexttt{real-test}}
\newcommand\hlloc[2][]{\tikz[overlay]\node[fill=mypurple, inner sep=0.5pt, anchor=text, rectangle,minimum height=3mm,#1] {#2};\phantom{#2}}
\newcommand\hlrep[2][]{\tikz[overlay]\node[fill=myyellow, inner sep=0.5pt, anchor=text, rectangle,minimum height=3mm,#1] {#2};\phantom{#2}}
\newcommand{\mycolorbox}[1]{\tikz{\draw[draw=drawgray, line width=0.75pt, fill=#1] (0, 0) rectangle (0.4, -0.25);}}

\DeclareMathOperator*{\softmax}{softmax}

\definecolor{mygreen}{rgb}{0.75, 0.95, 0.75}
\definecolor{myyellow}{rgb}{0.94,0.9,0.66}
\definecolor{mypurple}{rgb}{0.77, 0.77, 1}
\definecolor{myred}{rgb}{1,0.8,0.8}
\definecolor{myblue}{rgb}{0.6, 0.85, 1}
\definecolor{myorange}{rgb}{1, 0.85, 0.73}

\definecolor{arrowgray}{gray}{0.3}
\definecolor{drawgray}{gray}{0.5}
\definecolor{fillgray}{gray}{0.9}

\theoremstyle{plain}

\theoremstyle{definition}

\theoremstyle{remark}

\makeatletter
\tikzset{
        hatch distance/.store in=\hatchdistance,
        hatch distance=5pt,
        hatch thickness/.store in=\hatchthickness,
        hatch thickness=5pt
        }
\pgfdeclarepatternformonly[\hatchdistance,\hatchthickness]{north east hatch}
    {\pgfqpoint{-1pt}{-1pt}}
    {\pgfqpoint{\hatchdistance}{\hatchdistance}}
    {\pgfpoint{\hatchdistance-1pt}{\hatchdistance-1pt}}%
    {
        \pgfsetcolor{\tikz@pattern@color}
        \pgfsetlinewidth{\hatchthickness}
        \pgfpathmoveto{\pgfqpoint{0pt}{0pt}}
        \pgfpathlineto{\pgfqpoint{\hatchdistance}{\hatchdistance}}
        \pgfusepath{stroke}
    }
\pgfdeclarepatternformonly[\hatchdistance,\hatchthickness]{north west hatch}
    {\pgfqpoint{-1pt}{\hatchdistance}}
    {\pgfqpoint{\hatchdistance}{-1pt}}
    {\pgfpoint{\hatchdistance-1pt}{\hatchdistance-1pt}}%
    {
        \pgfsetcolor{\tikz@pattern@color}
        \pgfsetlinewidth{\hatchthickness}
        \pgfpathmoveto{\pgfqpoint{0pt}{\hatchdistance}}
        \pgfpathlineto{\pgfqpoint{\hatchdistance}{0pt}}
        \pgfusepath{stroke}
    }
\makeatother

\usepackage[textsize=tiny]{todonotes}

\icmltitlerunning{On Distribution Shift in Learning-based Bug Detectors}

\begin{document}

\twocolumn[
\icmltitle{On Distribution Shift in Learning-based Bug Detectors}



\icmlsetsymbol{equal}{*}

\begin{icmlauthorlist}
\icmlauthor{Jingxuan He}{eth}
\icmlauthor{Luca Beurer-Kellner}{eth}
\icmlauthor{Martin Vechev}{eth}
\end{icmlauthorlist}

\icmlaffiliation{eth}{Department of Computer Science, ETH Zurich, Switzerland}

\icmlcorrespondingauthor{Jingxuan He}{jingxuan.he@inf.ethz.ch}

\icmlkeywords{Machine Learning, ICML}

\vskip 0.3in
]



\printAffiliationsAndNotice{}  


\lstset{
  frame=l,
  language=Python,
  basicstyle=\fontsize{8}{9}\ttfamily,
  numbers=none,
  numberstyle=\tiny\color{gray},
  keywordstyle=\bfseries\color{blue},
  commentstyle=\color{gray}\ttfamily,
  keywordstyle=\textbf,
  escapeinside={(*@}{@*)},
  frame=none,
  upquote=true
  literate={``}{\textquotedblleft}1,
}

\setcounter{footnote}{1}
\begin{abstract}
  Deep learning has recently achieved initial success in program analysis tasks such as bug detection. Lacking real bugs, most existing works construct training and test data by injecting synthetic bugs into correct programs. Despite achieving high test accuracy (e.g., \textgreater90\%), the resulting bug detectors are found to be surprisingly unusable in practice, i.e., \textless10\% precision when used to scan real software repositories. In this work, we argue that this massive performance difference is caused by a distribution shift, i.e., a fundamental mismatch between the real bug distribution and the synthetic bug distribution used to train and evaluate the detectors. To address this key challenge, we propose to train a bug detector in two phases, first on a synthetic bug distribution to adapt the model to the bug detection domain, and then on a real bug distribution to drive the model towards the real distribution. During these two phases, we leverage a multi-task hierarchy, focal loss, and contrastive learning to further boost performance. We evaluate our approach extensively on three widely studied bug types, for which we construct new datasets carefully designed to capture the real bug distribution. The results demonstrate that our approach is practically effective and successfully mitigates the distribution shift: our learned detectors are highly performant on both our test set and the latest version of open source repositories. Our code, datasets, and models are publicly available at \url{https://github.com/eth-sri/learning-real-bug-detector}.
\end{abstract}

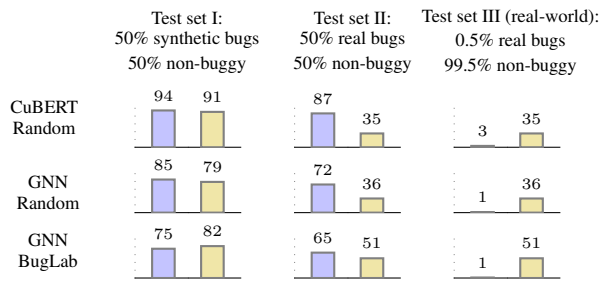
\begin{figure}[!t]
  \centering
  \begin{tikzpicture}[
    myarr/.style={line width=0.7mm, draw=arrowgray, -{Triangle[length=1.4mm,width=1.8mm]}, shorten >=2mm, shorten <=2mm},
  ]
    \begin{groupplot}[
      group style={group size=3 by 3, horizontal sep=20pt, vertical sep=10pt},
      width=3cm, height=2.1cm,
      ybar=10pt, /pgf/bar width=0.3cm,
      axis x line*=bottom, axis y line*=left, y axis line style={opacity=0},
      xtick={0}, xticklabels={}, yticklabels={}, tickwidth=0pt,
      ymin=0, ymax=100,
      every node near coord/.append style={font=\tiny},
      nodes near coords={\pgfmathprintnumber[precision=0]{\pgfplotspointmeta}},
    ]
      \node at (-1.2, 0.4) {\scriptsize \shortstack[c]{\cubert{}\\Random}};
      \node at (-1.15, -0.6) {\scriptsize \shortstack[c]{\gnn{}\\Random}};
      \node at (-1.15, -1.4) {\scriptsize \shortstack[c]{\gnn{}\\\buglab{}}};

      \node at (0.7, 1.4) {\scriptsize \shortstack[c]{Test set I:\\50\% synthetic bugs\\50\% non-buggy}};
      \node at (2.9, 1.4) {\scriptsize \shortstack[c]{Test set II:\\50\% real bugs\\50\% non-buggy}};
      \node at (5.0, 1.4) {\scriptsize \shortstack[c]{Test set III (real-world):\\0.5\% real bugs\\99.5\% non-buggy}};

      \nextgroupplot
        \addplot [draw=drawgray, fill=mypurple, line width=0.8pt] coordinates {(0, 94.43)};
        \addplot [draw=drawgray, fill=myyellow, line width=0.8pt] coordinates {(0, 90.77)};

      \nextgroupplot
        \addplot [draw=drawgray, fill=mypurple, line width=0.8pt] coordinates {(0, 86.86)};
        \addplot [draw=drawgray, fill=myyellow, line width=0.8pt] coordinates {(0, 35.42)};

      \nextgroupplot
        \addplot [draw=drawgray, fill=mypurple, line width=0.8pt] coordinates {(0, 3.43)};
        \addplot [draw=drawgray, fill=myyellow, line width=0.8pt] coordinates {(0, 35.42)};

      \nextgroupplot
        \addplot [draw=drawgray, fill=mypurple, line width=0.8pt] coordinates {(0, 84.94)};
        \addplot [draw=drawgray, fill=myyellow, line width=0.8pt] coordinates {(0, 78.87)};

      \nextgroupplot
        \addplot [draw=drawgray, fill=mypurple, line width=0.8pt] coordinates {(0, 72.19)};
        \addplot [draw=drawgray, fill=myyellow, line width=0.8pt] coordinates {(0, 36.31)};

      \nextgroupplot
        \addplot [draw=drawgray, fill=mypurple, line width=0.8pt] coordinates {(0, 1.31)};
        \addplot [draw=drawgray, fill=myyellow, line width=0.8pt] coordinates {(0, 36.31)};

      \nextgroupplot
        \addplot [draw=drawgray, fill=mypurple, line width=0.8pt] coordinates {(0, 75.14)};
        \addplot [draw=drawgray, fill=myyellow, line width=0.8pt] coordinates {(0, 81.85)};

      \nextgroupplot
        \addplot [draw=drawgray, fill=mypurple, line width=0.8pt] coordinates {(0, 65.13)};
        \addplot [draw=drawgray, fill=myyellow, line width=0.8pt] coordinates {(0, 50.60)};

      \nextgroupplot
        \addplot [draw=drawgray, fill=mypurple, line width=0.8pt] coordinates {(0, 1.16)};
        \addplot [draw=drawgray, fill=myyellow, line width=0.8pt] coordinates {(0, 50.60)};
    \end{groupplot}
  \end{tikzpicture}
  \vspace{-2mm}
  \caption{Performance of variable misuse classifiers are drastically reduced from synthetic test sets to a real-world test set. \mycolorbox{mypurple} and \mycolorbox{myyellow} are precision and recall, respectively.}
  \label{figure:shift}
\end{figure}
\section{Introduction}
\label{sec:intro}

The increasing amount of open source programs and advances in neural code models have stimulated the initial success of deep learning-based bug detectors~\cite{DBLP:conf/iclr/VasicKMBS19,DBLP:conf/iclr/HellendoornSSMB20,pybuglab,DBLP:conf/icml/KanadeMBS20,plur}. These detectors can discover hard-to-spot bugs such as variable misuses~\cite{DBLP:conf/iclr/AllamanisBK18} and wrong binary operators~\cite{DBLP:journals/pacmpl/PradelS18}, issues that greatly impair software reliability~\cite{DBLP:journals/pacmpl/RiceAJJPA17,DBLP:conf/msr/KarampatsisS20} and cannot be handled by traditional formal reasoning techniques.

Lacking real bugs, existing works build training sets by injecting few synthetic bugs, \eg, one~\cite{DBLP:conf/icml/KanadeMBS20}, three~\cite{DBLP:conf/iclr/HellendoornSSMB20}, or five~\cite{pybuglab}, into each correct program. The learned bug detectors then achieve high accuracy on test sets created in the same way as the training set. However, when scanning real-world software repositories, these detectors were found to be highly imprecise and practically ineffective, achieving only 2\% precision \cite{pybuglab} or \textless10\% precision \cite{DBLP:conf/pldi/HeLRV21}. The key question then is: what is the root cause for this massive drop in performance?

\paragraph{Unveiling Distribution Shifts}
We argue that the root cause is \emph{distribution shift}~\cite{DBLP:conf/icml/KohSMXZBHYPGLDS21}, a fundamental mismatch between the real bug distribution found in public code repositories and the synthetic bug distribution used to train and evaluate existing detectors. Concretely, real bugs are known to be different from synthetic ones~\cite{DBLP:conf/icml/YasunagaL21}, and further, correct programs outnumber buggy ones in practice, \eg, around 1:2000 as reported in~\cite{DBLP:conf/msr/KarampatsisS20}. This means that the real bug distribution inherently exhibits extreme data imbalance.

\cref{figure:shift} reproduces the performance drop, showing that existing detectors indeed fail to capture these two key factors. As in~\cite{DBLP:conf/icml/KanadeMBS20}, we fine-tune a classifier based on \cubert{} for variable misuse bugs, using a balanced dataset with randomly injected synthetic bugs. Non-surprisingly, the fine-tuned model is close-to-perfect on test set I created in the same way as the fine-tuning set (top-left of \cref{figure:shift}). Then, we replace the synthetic bugs in test set I with real bugs extracted from GitHub to create test set II (top-mid of \cref{figure:shift}). The precision and recall drop by 7\% and 56\%, respectively, meaning that the model is significantly worse at finding real bugs. Next, we evaluate the classifier on test set III created by adding a large amount of non-buggy code to test set II so to mimic the real-world data imbalance. The model achieves a precision of only 3\% (top-right of \cref{figure:shift}). A similar performance loss occurs with graph neural networks (\gnn{}s) \cite{DBLP:conf/iclr/AllamanisBK18}, trained on either the dataset for fine-tuning the previous \cubert{} model (mid row of \cref{figure:shift}) or another balanced dataset where synthetic bugs are injected by \buglab{}~\cite{pybuglab}, a learned bug selector (bottom row of \cref{figure:shift}).

\paragraph{This Work: Alleviating Distribution Shifts}
In this work, we aim to alleviate such a distribution shift and learn bug detectors capturing the real bug distribution. To achieve this goal, we propose to train bug detectors in two phases: (1) on a balanced dataset with synthetic bugs, similarly to existing works, and then (2) on a dataset that captures data imbalance and contains a small number of real bugs, possible to be extracted from GitHub commits~\cite{pybuglab} or industry bug archives~\cite{DBLP:journals/pacmpl/RiceAJJPA17}. In the first phase, the model deals with a relatively easier and larger training dataset. It quickly learns relevant features and captures the synthetic bug distribution. The second training dataset is significantly more difficult to learn from due to only a small number of positive samples and the extreme data imbalance. However, with the warm-up from the first training phase, the model can catch new learning signals and adapt to the real bug distribution. Such a two-phase learning process resembles the pre-training and fine-tuning scheme of large language models~\cite{DBLP:conf/naacl/DevlinCLT19} and self-supervised learning~\cite{DBLP:journals/pami/JingT21}. The two phases are indeed both necessary: as we show in \cref{sec:eval}, without either of the two phases or mixing them into one, the learned detectors achieve sub-optimal bug detection performance.

To boost performance, our learning framework also leverages additional components such as task hierarchy~\cite{DBLP:conf/acl/SogaardG16}, focal loss~\cite{DBLP:conf/iccv/LinGGHD17}, and a contrastive loss term to differentiate buggy/non-buggy pairs.

\paragraph{Datasets, Evaluation, and Effectiveness}
In our work, we construct datasets capturing the real bug distribution. To the best of our knowledge, these datasets are among the ones containing the largest number of real bugs (\eg, 1.7x of PyPIBugs \cite{pybuglab}) and are the first to capture data imbalance, for the bug types we handle. We use half of these datasets for our second phase of training and the other two quarters for validation and testing, respectively. Our extensive evaluation shows that our method is practically effective: it yields highly performant bug detectors that achieve matched precision (or the precision gap is greatly reduced) on our constructed test set and the latest version of open source repositories. This demonstrates that our approach successfully mitigates the challenge of distribution shift and our dataset is suitable for evaluating practical usefulness of bug detectors.

\section{Background}
\label{sec:background}
We now define the bug types our work handles and describe the state-of-the-art pointer models for detecting them.

\begin{figure*}
  \vspace{-2mm}
  \centering
  \begin{subfigure}[b]{0.32\textwidth}
    \centering
    \begin{lstlisting}[basicstyle=\scriptsize\ttfamily]
def compute_area(width, (*@\hlrep{height}@*)):
  return width * (*@\hlloc{width}@*)
    \end{lstlisting}
  \end{subfigure}\hfill
  \begin{subfigure}[b]{0.32\textwidth}
    \centering
    \begin{lstlisting}[basicstyle=\scriptsize\ttfamily]
+ - (*@\hlrep{*}@*) /
def compute_area(width, height):
  return width (*@\hlloc{+}@*) height
    \end{lstlisting}
  \end{subfigure}\hfill
  \begin{subfigure}[b]{0.32\textwidth}
    \begin{lstlisting}[basicstyle=\scriptsize\ttfamily]
def buy_with(account):
  return lib.withdraw((*@\hlloc{120.0}@*), (*@\hlrep{account}@*))
    \end{lstlisting}
  \end{subfigure}
  \vspace{-3mm}
  \caption{Example bugs handled by our work (left: \varmisuse{}, middle: \wrongbinop{}, right: \argswap{}). The bug location and the repair token are marked as \mycolorbox{mypurple} and \mycolorbox{myyellow}, respectively.}
  \label{fig:token-based-bugs}
\end{figure*}
\paragraph{Detecting Token-based Bugs}
We focus on token-based bugs caused by misuses of one or a few program tokens. One example is \varmisuse{} where a variable use is wrong and should be replaced by another variable defined in the same scope. Formally, we model a program $p$ as a sequence of $n$ tokens $T=\langle t_1, t_2, \ldots, t_n \rangle$. Fully handling a specific type of token-based bug in $p$ involves three tasks:
\begin{itemize}[leftmargin=*]
  \item \emph{Classification}: classify if $p$ is buggy or not.
  \item \emph{Localization}: if $p$ is buggy, locate the bug.
  \item \emph{Repair}: if $p$ is buggy, repair the located bug.
\end{itemize}
Note that these three tasks form a dependency where the later task depends on the prior tasks. To complete the three tasks, we first extract $Loc \subseteq T$, a set of candidate tokens from $T$ where a bug can be located. If $Loc = \varnothing$, $p$ is non-buggy, Otherwise, we say that $p$ is \emph{eligible} for bug detection and proceed with the classification task. We assign a value in $\{\pm 1\}$ to $p$, where $-1$ (\resp, $1$) means that $p$ is non-buggy (\resp, buggy). If $1$ is assigned to $p$, we continue with the localization and repair tasks. For localization, we identify a bug location token $t_{loc} \in Loc$. For repair, we apply simple rules (see \cref{sec:details}) to extract $Rep$, a set of candidate tokens that can be used to repair the bug located at $t_{loc}$ and find a token $t_{rep} \in Rep$ as the final repair token. $Loc$ and $Rep$ are defined based on the specific type of bug to detect. For example with \varmisuse{}, $Loc$ is the set of variable uses in $p$ and $Rep$ is the set of variables defined in the scope of the wrong variable use. The above definition is general and applies to the three popular types of token-based bugs handled in this work: \varmisuse{}, wrong binary operator (\wrongbinop{}), and argument swapping (\argswap{}). These bugs were initially studied in~\cite{DBLP:conf/iclr/AllamanisBK18,DBLP:journals/pacmpl/PradelS18}. We provide examples of those bugs in \cref{fig:token-based-bugs} and describe them in detail in \cref{sec:details}.

\paragraph{Existing Pointer Models for Token-based Bugs}
State-of-the-art networks for handling token-based bugs follow the design of \emph{pointer models}~\cite{DBLP:conf/iclr/VasicKMBS19}, which identify bug locations and repair tokens based on predicted pointer vectors. Given the program tokens $T$, pointer models first apply an embedding method $\phi$ to convert each token $t_i$ into an $m$-dimensional feature vector $h_i \in \mathbb{R}^m$:
\begin{equation*}
  [h_1, \ldots, h_n] = \phi(\langle t_1, \ldots, t_n \rangle).
\end{equation*}

Existing works instantiate $\phi$ as GNNs and GREATs~\cite{DBLP:conf/iclr/HellendoornSSMB20}, LSTMs~\cite{DBLP:conf/iclr/VasicKMBS19}, or BERT~\cite{DBLP:conf/icml/KanadeMBS20}. Then, a feedforward network $\pi^{loc}$ is applied on the feature vectors to obtain a probability vector $P^{loc} = [p^{loc}_1, \ldots, p^{loc}_n]$ pointing to the bug location. The repair probabilities $P^{rep} = [p^{rep}_1, \ldots, p^{rep}_l]$ are computed in a similar way with another feedforward network $\pi^{rep}$. We omit the steps for computing $P^{loc}$ and $P^{rep}$ for brevity and elaborate on them in \cref{sec:compute-repair}.

Importantly, in existing pointer models, classification is done \emph{jointly} with localization. That is, each program has a special \nobug{} location (typically the first token). When the localization result points to \nobug{}, the classification result is $-1$. Otherwise, the localization result points to a bug location and the classification result is $1$.

\paragraph{Training Existing Pointer Models}
For training, two masks, $C^{loc}$ and $C^{rep}$, are required as ground truth labels. $C^{loc}$ sets the index of the correct bug location as $1$ and other indices to $0$. Similarly, $C^{rep}$ sets the indices of the correct repair tokens as $1$ and other indices to $0$. The localization and repair losses are:
\begin{align*}
  L^{loc} &= -\sum\nolimits_{i} p^{loc}_i \times C^{loc}[i], \\ 
  L^{rep} &= -\sum\nolimits_{i} p^{rep}_i \times C^{rep}[i].
\end{align*}
The additive loss $L = L^{loc} + L^{rep}$ is optimized. In \cref{sec:method}, we introduce additional loss terms to $L$.

\section{Learning Distribution-Aware Bug Detectors}
\label{sec:method}

Building on the pointer models discussed in \cref{sec:background}, we now present our framework for learning bug detectors capable of capturing the real bug distribution.

\subsection{Network Architecture with Multi-Task Hierarchy}
We first describe architectural changes to the pointer model.

\paragraph{Adding Classification Head}
In our early experiments, we found that a drawback of pointer models is mixing the classification and localization results in one pointer vector. As a result, the model can be confused by the two tasks. We propose to perform the two tasks individually by adding a \emph{binary classification head}: We treat the first token $t_1$ as the classification token $t_{[cls]}$ and apply a feedforward network $\pi^{cls}: \mathbb{R}^{m} \rightarrow \mathbb{R}^2$ over its feature vector $h_{[cls]}$ to compute the classification probabilities $p^{cls}_{-1}$ and $p^{cls}_{1}$:
\begin{equation*}
  [p^{cls}_{-1},\; p^{cls}_{1}] = \softmax(\pi^{cls}(h_{[cls]})).
\end{equation*}

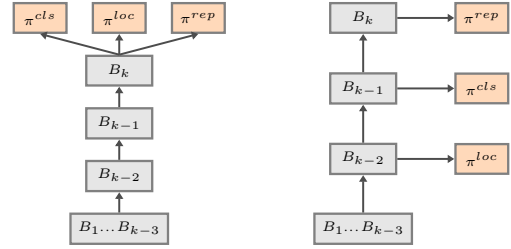
\begin{figure}[!t]
  \centering
  \resizebox{0.8\columnwidth}{!}{\begin{tikzpicture}[
    block/.style={rectangle, draw=drawgray, fill=fillgray, line width=1pt, minimum width=1cm, minimum height=0.45cm},
    head/.style={rectangle, draw=drawgray, fill=myorange, line width=1pt, minimum width=0.8cm, minimum height=0.45cm},
    myarr/.style={->, line width=0.3mm, draw=arrowgray, -{Triangle[length=1mm,width=1.2mm]}},
  ]
    \begin{scope}[shift={(0, 0)}]
      \node (bk3) [block] at (0, 0) {\tiny $B_1$\ldots$B_{k-3}$};
      \node (bk2) [block] at (0, 0.8) {\tiny $B_{k-2}$};
      \node (bk1) [block] at (0, 1.6) {\tiny $B_{k-1}$};
      \node (bk) [block] at (0, 2.4) {\tiny $B_k$};
      \node (cls) [head] at (-1.2, 3.2) {\tiny $\pi^{cls}$};
      \node (loc) [head] at (0, 3.2) {\tiny $\pi^{loc}$};
      \node (rep) [head] at (1.2, 3.2) {\tiny $\pi^{rep}$};

      \draw[myarr] (bk3.north) -- (bk2.south);
      \draw[myarr] (bk2.north) -- (bk1.south);
      \draw[myarr] (bk1.north) -- (bk.south);
      \draw[myarr] (bk.north) -- (cls.south);
      \draw[myarr] (bk.north) -- (loc.south);
      \draw[myarr] (bk.north) -- (rep.south);
    \end{scope}

    \begin{scope}[shift={(3.7, 0)}]
      \node (bk3) [block] at (0, 0) {\tiny $B_1$\ldots$B_{k-3}$};
      \node (dk2) [head] at (1.8, 1.06) {\tiny $\pi^{loc}$};
      \node (bk2) [block, fill=fillgray] at (0, 1.06) {\tiny $B_{k-2}$};
      \node (dk1) [head] at (1.8, 2.13) {\tiny $\pi^{cls}$};
      \node (bk1) [block, fill=fillgray] at (0, 2.13) {\tiny $B_{k-1}$};
      \node (dk) [head] at (1.8, 3.2) {\tiny $\pi^{rep}$};
      \node (bk) [block, fill=fillgray] at (0, 3.2) {\tiny $B_k$};

      \draw[myarr] (bk3.north) -- (bk2.south);
      \draw[myarr] (bk2.north) -- (bk1.south);
      \draw[myarr] (bk1.north) -- (bk.south);
      \draw[myarr] (bk2.east) -- (dk2.west);
      \draw[myarr] (bk1.east) -- (dk1.west);
      \draw[myarr] (bk.east) -- (dk.west);
    \end{scope}
  \end{tikzpicture}}
  \vspace{-1mm}
  \caption{Standard feature sharing (left) \vs{} an example of our task hierarchy (right).}
  \label{figure:hierarchy}
\end{figure}
\paragraph{Task Hierarchy}
To exploit the inter-dependence of the $cls$, $loc$, and $rep$ tasks, we formulate a \emph{task hierarchy} for the pointer model. This allows the corresponding components to reinforce each other and improve overall performance. Task hierarchies are a popular multi-task learning technique~\cite{DBLP:journals/corr/ZhangY17aa} and are effective in natural language~\cite{DBLP:conf/acl/SogaardG16} and computer vision~\cite{DBLP:conf/eccv/GuoHHYF18}. To the best of our knowledge, this work is the first to apply a task hierarchy on code tasks.

Using a task hierarchy, we process each task one by one following a specific order instead of addressing all tasks in the same layer. Formally, to encode a task hierarchy, we consider the feature embedding function $\phi$ to consist of $k$ feature transformation layers $B_1, \ldots, B_k$, which is a standard design of existing pointer models~\cite{DBLP:conf/iclr/HellendoornSSMB20,pybuglab,DBLP:conf/icml/KanadeMBS20}:
\begin{equation*}
  \phi(T) =  (B_k \circ B_{k-1} \circ \ldots \circ B_1)(T).
\end{equation*}
We order our tasks $cls$, $loc$ and $rep$, and apply their feedforward networks separately on the last three feature transformation layers. \cref{figure:hierarchy} shows a task hierarchy with the order $[loc, cls, rep]$ and compares it with feature sharing.

\subsection{Imbalanced Classification with Focal Loss}
To handle the extreme data imbalance in practical bug classification, we leverage the focal loss from imbalanced learning in computer vision and natural language processing~\cite{DBLP:conf/iccv/LinGGHD17,DBLP:conf/acl/LiSMLWL20}. Focal loss is defined as:
\begin{equation*}
  L^{cls} = \text{FL}([p^{cls}_{-1},\; p^{cls}_{1}], y) = -(1-p^{cls}_y)^{\gamma}\log(p^{cls}_y),
\end{equation*}
where $y$ is the ground truth label. Compared with the standard cross entropy loss, focal loss adds an adjusting factor $(1-p^{cls}_y)^{\gamma}$ serving as an importance weight for the current sample. When the model has high confidence with large $p^{cls}_y$, the adjusting factor becomes exponentially small. This helps the model put less attention on the large volume of easy, negative samples and focus on hard samples which the model is unsure about. $\gamma$ is a tunable parameter that we set to $2$ according to \cite{DBLP:conf/iccv/LinGGHD17}.

\subsection{Two-phase Training}
\label{sec:two-step}

Bug detectors are ideally trained on a dataset containing a large number of real bugs as encountered in practice. However, since bugs are scarce, the ideal dataset does not exist and is hard to obtain with either manual or automatic approaches~\cite{DBLP:conf/pldi/HeLRV21}. Only a small number of real bugs can be extracted from GitHub commits~\cite{pybuglab,DBLP:conf/msr/KarampatsisS20} or industry bug archives~\cite{DBLP:journals/pacmpl/RiceAJJPA17}, which are not sufficient for data intensive deep models. Our two-phase training overcomes this dilemma by utilizing both synthetic and real bugs.

\paragraph{Phase 1: Training with large amounts of synthetic bugs}
In the first phase, we train the model using a dataset containing a large number of synthetic bugs and their correct version. Even though learning from such a dataset does not yield a final model that captures the real bug distribution, it drives the model to learn relevant features for bug detection and paves the way for the second phase. We create this dataset following existing work~\cite{DBLP:conf/icml/KanadeMBS20}: we obtain a large set of open source programs $P$, inject synthetic bugs into each program $p \in P$, and create a buggy program $p'$, which results in an $1:1$ balanced dataset.

Since $p$ and $p'$ only differ in one or a few tokens, the model sometimes struggles to distinguish between the two, which impairs classification performance. We alleviate this issue by forcing the model to produce distant classification feature embeddings $h_{[cls]}$ for $p$ and $h_{[cls]}'$ for $p'$. This is achieved by a contrastive loss term measuring the cosine similarity of $h_{[cls]}$ and $h_{[cls]}'$:
\begin{equation*}
  L^{contrastive} = \cos(h_{[cls]}, h'_{[cls]}).
\end{equation*}

The final loss in the first training phase is the addition of four loss terms: $L^{cls}$, $L^{loc}$, $L^{rep}$, and $\beta L^{contrastive}$, where $\beta$ is the tunable weight of the contrastive loss.

\paragraph{Phase 2: Training with real bugs and data imbalance}
The second training phase provides additional supervision to drive the model trained in phase 1 to the real bug distribution. To achieve this, we leverage a training dataset that mimics the real bug distribution. The dataset contains a small number of real bugs (typically hundreds) extracted from GitHub commits, which helps the model to adapt from synthetic bugs to real ones. Moreover, we add a large amount of non-buggy samples (\eg, hundreds of thousands) to mimic the real data imbalance. For more details on how we construct this dataset, please see \cref{sec:impl}. The loss for the second training phase is the sum of the three task losses $L^{cls}$, $L^{loc}$, and $L^{rep}$.

\section{Implementation and Dataset Construction}
\label{sec:impl}

In this section, we discuss the implementation of our learning framework and our dataset construction procedure.

\paragraph{Leveraging \cubert{}}
We implement our framework on top of \cubert{}~\cite{DBLP:conf/icml/KanadeMBS20}, a BERT-like model~\cite{DBLP:conf/naacl/DevlinCLT19} pretrained on source code. Still, our framework is general and can be applied to any existing pointer models (as we show in \cref{sec:eval-two-phase}, our two-phase training brings improvement to the \gnn{} model in~\cite{pybuglab}). We chose \cubert{} mainly because, according to~\cite{plur}, it achieves top-level performance on many programming tasks, including detecting synthetic \varmisuse{} and \wrongbinop{} bugs. Moreover, the reproducibility package provided by the \cubert{} authors is of high quality and easy to extend. We present implementation details in \cref{sec:details}.

\begin{figure}[!t]
  \centering
  \resizebox{0.85\columnwidth}{!}{\begin{tikzpicture}[
    myarr/.style={line width=0.7mm, draw=arrowgray, -{Triangle[length=1.4mm,width=1.8mm]}, shorten >=2mm, shorten <=2mm},
  ]
    \begin{scope}[shift={(0.3, -0.4)}]
      \filldraw[draw=drawgray, fill=fillgray, line width=1pt] (0, 0) rectangle (1, -2);
      \node at (0.5, -2.4) {\scriptsize \shortstack[c]{Open source\\repositories}};
    \end{scope}

    \begin{scope}[shift={(2, -1.6)}]
      \node at (0.25, 0.45) {\scriptsize \shortstack[c]{Real bugs\\found?}};
      \draw[line width=0.7mm, draw=arrowgray] (-0.3, 0) -- (0.8, 0);
      \node at (1.5, 0.75) {\scriptsize No};
      \draw[myarr] (0.6, -0.12) -- (1.8, 0.6);
      \draw[myarr] (0.6, 0.12) -- (1.8, -0.6);
      \node at (1.5, -0.75) {\scriptsize Yes};
    \end{scope}

    \begin{scope}[shift={(4, 0)}]
      \draw[draw=drawgray, line width=1pt, pattern color=mypurple, pattern=north west hatch, hatch distance=7pt, hatch thickness=2pt] (0, 0) rectangle (1.2, -1.3);

      \node at (1.8, -0.45) {\scriptsize \shortstack[c]{Inject\\bugs}};
      \draw[myarr] (1.32, -0.9) -- (2.32, -0.9);

      \begin{scope}[shift={(2.4, 0)}]
        \draw[draw=drawgray, line width=1pt, pattern color=myyellow, pattern=north west hatch, hatch distance=7pt, hatch thickness=2.01pt] (0, 0) rectangle (1.2, -1.3);
      \end{scope}
      \node at (1.82, -1.5) {\scriptsize \texttt{syn-train}};
    \end{scope}

    \begin{scope}[shift={(4.2, -2)}]
      \draw[draw=drawgray, line width=1pt, pattern color=myyellow, pattern=north east hatch, hatch distance=7pt, hatch thickness=2.01pt] (0, 0) rectangle (0.3, -0.5);
      \draw[draw=drawgray, line width=1pt, pattern color=mypurple, pattern=north east hatch, hatch distance=7pt, hatch thickness=2pt] (0.5, 0) rectangle (1.6, -0.5);
      \node at (2.6, -0.25) {\scriptsize \texttt{real-train}};

      \draw[draw=drawgray, line width=1pt, pattern color=myyellow, pattern=north east hatch, hatch distance=7pt, hatch thickness=2.01pt] (0, -0.6) rectangle (0.3, -0.85);
      \draw[draw=drawgray, line width=1pt, pattern color=mypurple, pattern=north east hatch, hatch distance=7pt, hatch thickness=2pt] (0.5, -0.6) rectangle (1.6, -0.85);
      \node at (2.46, -0.7) {\scriptsize \texttt{real-val}};

      \draw[draw=drawgray, line width=1pt, pattern color=myyellow, pattern=north east hatch, hatch distance=7pt, hatch thickness=2.01pt] (0, -0.95) rectangle (0.3, -1.2);
      \draw[draw=drawgray, line width=1pt, pattern color=mypurple, pattern=north east hatch, hatch distance=7pt, hatch thickness=2pt] (0.5, -0.95) rectangle (1.6, -1.2);
      \node at (2.53, -1.05) {\scriptsize \texttt{real-test}};
    \end{scope}
  \end{tikzpicture}}
  \vspace{-1mm}
  \caption{Our data construction process. For the precise sizes the datasets, refer to \cref{table:dataset}.}
  \label{figure:dataset}
\end{figure}
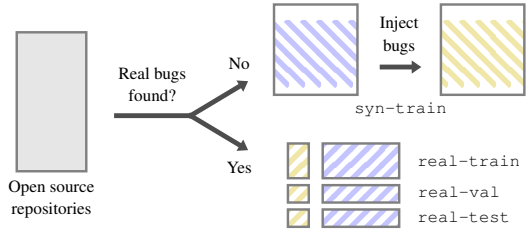

\begin{table*}[!t]
  \small
  \centering
  \def\arraystretch{1.1}
  \setlength\tabcolsep{6pt}
  \caption{The statistics of our constructed dataset.}
  \vspace{-2mm}
  \label{table:dataset}
  \resizebox{\textwidth}{!}{\begin{tabular}{@{}llcrlcrlcrlcr@{}}
    \toprule
    \multirow{3}{*}{Bug Type} & \multicolumn{3}{c}{\strain{}} & \multicolumn{3}{c}{\rtrain{}} & \multicolumn{3}{c}{\rval{}} & \multicolumn{3}{c}{\rtest{}} \\
    \cmidrule(lr){2-4} \cmidrule(lr){5-7} \cmidrule(lr){8-10} \cmidrule(l){11-13}
    & repo & buggy & non-buggy & repo & buggy & non-buggy & repo & buggy & non-buggy & repo & buggy & non-buggy \\
    \midrule
    \varmisuse{} & 2, 654 & 147, 409 & 147, 409 & 339 & 626 & 118, 888 & 169 & 347 & 63, 703 & 170 & 336 & 61, 539 \\
    \wrongbinop{} & 4, 944 & 150, 825 & 150, 825 & 368 & 872 & 73, 015 & 184 & 356 & 20, 341 & 185 & 491 & 41, 303 \\
    \argswap{} & 2, 009 & 157, 530 & 157, 530 & 372 & 469 & 82, 442 & 186 & 218 & 40, 305 & 185 & 246 & 48, 473 \\
    \bottomrule
  \end{tabular}}
\end{table*}
\paragraph{Dataset Construction}
We focus on Python code. \cref{figure:dataset} shows our dataset construction process. Careful deduplication was applied throughout the entire process~\cite{DBLP:conf/oopsla/Allamanis19}. After construction, we obtain a balanced dataset with synthetic bugs, called \strain{}, used for the first training phase. Moreover, we obtain an imbalanced dataset with real bugs, which is randomly split into \rtrain{} (used for the second training phase), \rval{} (used as the validation set), and \rtest{} (used as the blind test set). The split ratio is 0.5:0.25:0.25. Instead of splitting by files, we split the dataset by repositories. This prevents distributing files from the same repositories into different splits and requires generalization across codebases~\cite{DBLP:conf/icml/KohSMXZBHYPGLDS21}. Note that we do not evaluate on synthetic bugs as it does not reflect the practical usage. The statistics of the constructed datasets are given in \cref{table:dataset}.

In \cref{figure:dataset}, we start the construction with a set of open source repositories (ETH Py150 Open~\cite{ethpy150open,DBLP:conf/oopsla/RaychevBV16} for \varmisuse{} and \wrongbinop{}, and on top of that 894 additional repositories for \argswap{} to collect enough real bugs). We go over the commit history of the repositories and extract real bugs that align with the bug-inducing rewrite rules of~\cite{pybuglab} applied to both versions of a changed file. The repositories are then split into two sets depending on whether any real bug is found. To construct \strain{}, we extract $\sim$150k eligible functions as non-buggy samples for each bug type from the repositories in which no real bugs were found (\tikz{\draw[draw=drawgray, line width=0.75pt, pattern color=mypurple, pattern=north west hatch, hatch distance=4.5pt, hatch thickness=1.401pt] (0, 0) rectangle (0.4, -0.25);}). Then, we inject bugs into \tikz{\draw[draw=drawgray, line width=0.75pt, pattern color=mypurple, pattern=north west hatch, hatch distance=4.5pt, hatch thickness=1.401pt] (0, 0) rectangle (0.4, -0.25);} to create synthetic buggy samples (\tikz{\draw[draw=drawgray, line width=0.75pt, pattern color=myyellow, pattern=north west hatch, hatch distance=4.5pt, hatch thickness=1.4pt] (0, 0) rectangle (0.4, -0.25);}). The bugs are selected uniformly at random among all bug candidates. We leave the use of more advanced learning methods for bug selection~\cite{DBLP:conf/sigsoft/PatraP21,DBLP:conf/icml/YasunagaL21} as future work.
To construct \rtrain{}, we combine the found real bugs (\tikz{\draw[draw=drawgray, line width=0.75pt, pattern color=myyellow, pattern=north east hatch, hatch distance=4.5pt, hatch thickness=1.4pt] (0, 0) rectangle (0.4, -0.25);}) and other eligible functions from the same repositories, which serve as non-buggy samples (\tikz{\draw[draw=drawgray, line width=0.75pt, pattern color=mypurple, pattern=north east hatch, hatch distance=4.5pt, hatch thickness=1.401pt] (0, 0) rectangle (0.4, -0.25);}). Since the number of eligible functions is significantly larger than the number of real bugs, real data imbalance is preserved. Finally, we perform random splitting to obtain \texttt{real-}datasets as discussed before. 

\section{Experimental Evaluation}
\label{sec:eval}

In this section, we present an extensive evaluation of our framework. We first describe our experimental setup.

\paragraph{Training and Model}
We perform training per bug type because the three bug types have different characteristics. We tried a number of training configurations and found that our full method with all the techniques described in \cref{sec:method} performed the best on the validation set. For \varmisuse{}, \wrongbinop{}, and \argswap{}, the best orders for the task hierarchy are $[cls, loc, rep]$, $[rep, loc, cls]$, and $[loc, cls, rep]$, respectively. The best $\beta$ for the contrastive loss are 0.5, 4, and 0.5, respectively. We provide training and model details in \cref{sec:details}.

\begin{table*}[!t]
  \centering
  \begin{minipage}[t]{\columnwidth}
    \small
\centering
\def\arraystretch{1.2}
\setlength\tabcolsep{7pt}
\caption{Changing training phases (\varmisuse{}).}
\vspace{-2mm}
\label{table:two-phase-varmisuse}
\resizebox{\columnwidth}{!}{\begin{tabular}{@{}llrlrlr@{}}
  \toprule
  \multirow{3}{*}{Method} & \multicolumn{2}{c}{\texttt{cls}} & \multicolumn{2}{c}{\texttt{cls-loc}} & \multicolumn{2}{c}{\hspace{2mm}\texttt{cls-loc-rep}} \\
  \cmidrule(lr){2-3} \cmidrule(lr){4-5} \cmidrule(l){6-7}
  & P & R & P & R & P & R \\
  \midrule
  Only Synthetic & 3.43 & 35.42 & 2.45 & 25.30 & 2.10 & 21.73 \\
  Only Real & 0 & 0 & 0 & 0 & 0 & 0 \\
  Mix & 5.66 & 24.11 & 4.61 & 19.64 & 4.19 & 17.86 \\
  Two Synthetic & 35.59 & 6.25 & 32.20 & 5.65 & 30.51 & 5.36 \\
  \midrule
  Our Full Method & \textbf{64.79} & 13.69 & \textbf{61.97} & 13.10 & 56.34 & 11.90 \\
  \bottomrule
\end{tabular}}
  \end{minipage}\hfill
  \begin{minipage}[t]{\columnwidth}
    \small
\centering
\def\arraystretch{1.2}
\setlength\tabcolsep{7pt}
\caption{Changing training phases (\wrongbinop{}).}
\vspace{-2mm}
\label{table:two-phase-wrongbinop}
\resizebox{\columnwidth}{!}{\begin{tabular}{@{}llrlrlr@{}}
  \toprule
  \multirow{3}{*}{Method} & \multicolumn{2}{c}{\texttt{cls}} & \multicolumn{2}{c}{\texttt{cls-loc}} & \multicolumn{2}{c}{\hspace{2mm}\texttt{cls-loc-rep}} \\
  \cmidrule(lr){2-3} \cmidrule(lr){4-5} \cmidrule(l){6-7}
  & P & R & P & R & P & R \\
  \midrule
  Only Synthetic & 9.69 & 49.08 & 8.93 & 45.21 & 8.09 & 40.94 \\
  Only Real & 47.74 & 25.87 & 45.49 & 24.64 & 42.11 & 22.81 \\
  Mix & 12.97 & 41.55 & 12.02 & 38.49 & 10.93 & 35.03 \\
  Two Synthetic & 26.85 & 5.91 & 25.00 & 5.50 & 21.30 & 4.68 \\
  \midrule
  Our Full Method & \textbf{52.30} & 43.99 & \textbf{51.09} & 42.97 & \textbf{49.64} & 41.75 \\
  \bottomrule
\end{tabular}}
  \end{minipage}
\end{table*}
\begin{figure*}[!t]
  \centering
  \begin{minipage}[t]{\columnwidth}
    \centering
\begin{tikzpicture}
  \begin{groupplot}[
    width=4.5cm, height=4cm,
    compat=newest,
    group style={group size=2 by 1, horizontal sep=35pt},
    title style={at={(0.5,-0.25)}, anchor=north, yshift=-10, font=\footnotesize},
    ticklabel style={font=\tiny},
    tick align=outside,
    major tick length=2pt,
    xtick = {0, 20, 40, 60, 80, 100},
    axis x line*=bottom,
    xmin=0.0,
    xmax=100,
    xlabel = {\scriptsize R\textsuperscript{\texttt{cls}}},
    xlabel style = {at={(0.5, -0.15)}},
    ylabel = {\scriptsize P\textsuperscript{\texttt{cls}}},
    ylabel style = {at={(-0.17, 0.5)}},
    ymin=0,
    ymax=101,
    ytick = {0, 20, 40, 60, 80, 100},
    axis y line*=left,
    legend style={
      nodes={scale=0.4, transform shape},
      at={(1.1,01)},
      draw=none,
    },
    legend cell align={left},
    legend image code/.code={
      \draw[line width=0.8pt] plot coordinates {
        (0cm, 0cm)
        (0.2cm, 0cm)
        (0.4cm, 0cm)
      };
    }, 
  ]

    \nextgroupplot[
      title={(a) \texttt{var-misuse}},
    ]
      \input{figures/rp/real/var-misuse}
    \nextgroupplot[
      title={(b) \texttt{wrong-binop}},
    ]
      \input{figures/rp/real/wrong-bin-op}
  
  \end{groupplot}
\end{tikzpicture}
\vspace{-6mm}
\caption{The effectiveness of our two-phase training demonstrated by precision-recall curves and AP.}
\label{fig:pr-ours}
  \end{minipage}\hfill
  \begin{minipage}[t]{\columnwidth}
    \centering
\begin{tikzpicture}
  \begin{groupplot}[
    width=4.5cm, height=3.5cm,
    compat=newest,
    group style={group size=2 by 1, horizontal sep=35pt},
    tick align=outside,
    major tick length=2.5pt,
    title style={at={(0.5,-0.37)}, anchor=north, yshift=-10, font=\footnotesize},
    ticklabel style={font=\tiny},
    axis x line*=bottom,
    xmin=1,
    xmax=200,
    xmode=log,
    log basis x=2,
    xlabel = {\scriptsize Non-buggy/buggy ratio},
    xlabel style = {at={(0.5, -0.24)}},
    ylabel = {\scriptsize Performance},
    ylabel style = {at={(-0.17, 0.5)}},
    ymin=0,
    ymax=100,
    ytick = {0, 25, 50, 75, 100},
    axis y line*=left,
    legend style={
      nodes={scale=0.75, transform shape},
      at={(-0.3,1.5)},
      anchor=north,
      legend columns=4,
      draw=none,
      align=center,
      /tikz/every even column/.append style={column sep=0.3cm},
    },
    legend image code/.code={
      \draw[line width=1.5pt] plot coordinates {
        (0cm, 0cm)
        (0.2cm, 0cm)
        (0.4cm, 0cm)
      };
    }, 
  ]
    \nextgroupplot[
      title={\shortstack[c]{(a) \texttt{\footnotesize var-misuse}}},
      xtick = {1, 2, 4, 8, 16, 32, 64, 190},
      xticklabels = {2\textsuperscript{0}, 2\textsuperscript{1}, 2\textsuperscript{2}, 2\textsuperscript{3}, 2\textsuperscript{4}, 2\textsuperscript{5}, 2\textsuperscript{6}, 190},
    ]
      \input{figures/skew/varmisuse}

    \nextgroupplot[
      title={\shortstack[c]{(b) \texttt{\footnotesize wrong-binop}}},
      xtick = {1, 2, 4, 8, 16, 32, 64, 100},
      xticklabels = {2\textsuperscript{0}, 2\textsuperscript{1}, 2\textsuperscript{2}, 2\textsuperscript{3}, 2\textsuperscript{4}, 2\textsuperscript{5}, 2\textsuperscript{6}, 84},
    ]
      \input{figures/skew/wrongbinop}

    \legend{$\text{AP}$, $\text{P}^{\texttt{cls}}$, $\text{R}^{\texttt{cls}}$, $\text{P}^{\texttt{cls-loc}}$, $\text{R}^{\texttt{cls-loc}}$, $\text{P}^{\texttt{cls-loc-rep}}$, $\text{R}^{\texttt{cls-loc-rep}}$}
  \end{groupplot}
\end{tikzpicture}
{\phantomsubcaption\label{fig:skew-varmisuse}}
{\phantomsubcaption\label{fig:skew-wrongbinop}}
\vspace{-6mm}
\caption{Model performance with various non-buggy/buggy ratios in the second training phase.}
\label{fig:skew}
  \end{minipage}
\end{figure*}
\begin{figure*}[!t]
  \centering
  \begin{tikzpicture}
    \begin{groupplot}[
      width=4.5cm, height=4cm,
      compat=newest,
      group style={group size=4 by 1, horizontal sep=40pt},
      tick align=outside,
      major tick length=2pt,
      title style={at={(0.5,-0.25)}, anchor=north, yshift=-10, font=\footnotesize},
      ticklabel style={font=\scriptsize},
      xtick = {0.8, 2, 4, 8, 16, 32, 64, 100},
      xticklabels = {0, 2\textsuperscript{1}, 2\textsuperscript{2}, 2\textsuperscript{3}, 2\textsuperscript{4}, 2\textsuperscript{5}, 2\textsuperscript{6}, 100},
      axis x line*=bottom,
      xmin=0.8,
      xmax=100,
      xmode=log,
      log basis x=2,
      axis x discontinuity=crunch,
      xlabel = {\scriptsize Percentage},
      xlabel style = {at={(0.5, -0.15)}},
      ylabel = {\scriptsize Performance},
      ylabel style = {at={(-0.17, 0.5)}},
      ymin=0,
      ymax=100,
      ytick = {0, 25, 50, 75, 100},
      axis y line*=left,
      legend style={
        nodes={scale=0.9, transform shape},
        at={(-1.7,1.3)},
        anchor=north,
        legend columns=-1,
        draw=none,
        align=center,
        /tikz/every even column/.append style={column sep=0.4cm},
      },
      legend image code/.code={
        \draw[line width=1.5pt] plot coordinates {
          (0cm, 0cm)
          (0.26cm, 0cm)
          (0.52cm, 0cm)
        };
      }, 
    ]
      \nextgroupplot[
        title={\shortstack[c]{(a) \texttt{\footnotesize synthetic-train}\\\texttt{\footnotesize var-misuse}}},
      ]
        \input{figures/synthetic/varmisuse}

      \nextgroupplot[
        title={\shortstack[c]{(d) \texttt{\footnotesize real-train}\\\texttt{\footnotesize var-misuse}}},
      ]
        \input{figures/real/varmisuse}

      \nextgroupplot[
        title={\shortstack[c]{(b) \texttt{\footnotesize synthetic-train}\\\texttt{\footnotesize wrong-binop}}},
      ]
        \input{figures/synthetic/wrongbinop}
  
      \nextgroupplot[
        title={\shortstack[c]{(e) \texttt{\footnotesize real-train}\\\texttt{\footnotesize wrong-binop}}},
      ]
        \input{figures/real/wrongbinop}

      \legend{$\text{AP}$, $\text{P}^{\texttt{cls}}$, $\text{R}^{\texttt{cls}}$, $\text{P}^{\texttt{cls-loc}}$, $\text{R}^{\texttt{cls-loc}}$, $\text{P}^{\texttt{cls-loc-rep}}$, $\text{R}^{\texttt{cls-loc-rep}}$}
    \end{groupplot}
  \end{tikzpicture}
  {\phantomsubcaption\label{fig:synthetic-varmisuse}}
  {\phantomsubcaption\label{fig:real-varmisuse}}
  {\phantomsubcaption\label{fig:synthetic-wrongbinop}}
  {\phantomsubcaption\label{fig:real-wrongbinop}}
  \vspace{-6mm}
  \caption{Model performance with subsampled \strain{} or \rtrain{}.}
  \label{fig:percent}
\end{figure*}
\paragraph{Testing and Metrics}
We perform testing on the \rtest{} dataset. For space reasons, we only discuss the testing results for \varmisuse{} and \wrongbinop{} in this section. We show the results for \argswap{} in \cref{sec:more-eval}.

Instead of accuracy~\cite{pybuglab,DBLP:conf/iclr/HellendoornSSMB20,DBLP:conf/iclr/VasicKMBS19}, we use precision and recall that are known to be better suited for data imbalance settings~\cite{wiki,saito2015precision}. They are computed per evaluation target \mytexttt{tgt} as follows:
\begin{equation*}
  \text{P}^{\texttt{tgt}} = \frac{tp^{\texttt{tgt}}}{tp^{\texttt{tgt}} + fp^{\texttt{tgt}}}, \quad\quad\quad \text{R}^{\texttt{tgt}} = \frac{tp^{\texttt{tgt}}}{\small \#\texttt{buggy}},
\end{equation*}
where {\small $\#\texttt{buggy}$} is the number of samples labeled as buggy. When $tp^{\texttt{tgt}} + fp^{\texttt{tgt}} = 0$, we assign $\text{P}^{\texttt{tgt}} = 0$. We consider three evaluation targets \mytexttt{cls}, \mytexttt{cls-loc}, and \mytexttt{cls-loc-rep}:
\begin{itemize}[leftmargin=*]
  \item \mytexttt{cls}: binary classification. $tp^{\texttt{cls}}$ means that the classification prediction and the ground truth are both buggy. A sample is an $fp^{\texttt{cls}}$ when the classification result is buggy but the ground truth is non-buggy.
  \item \mytexttt{cls-loc}: joint classification and localization. A sample is a $tp^{\texttt{cls-loc}}$ when it is a $tp^{\texttt{cls}}$ and the localization token is correctly predicted. A sample is an $fp^{\texttt{cls-loc}}$ when it is a $fp^{\texttt{cls}}$ or a $tp^{\texttt{cls}}$ with wrong localization result.
  \item \mytexttt{cls-loc-rep}: joint classification, localization, and repair. A sample is a $tp^{\texttt{cls-loc-rep}}$ when it is a $tp^{\texttt{cls-loc}}$ and the repair token is correctly predicted. A sample is an $fp^{\texttt{cls-loc-rep}}$ when it is a $fp^{\texttt{cls-loc}}$ or a $tp^{\texttt{cls-loc}}$ with wrong repair result.
\end{itemize}
The dependency between classification, localization, and repair determines how a bug detector is used in practice. That is, the users will look at the localization result only when the classification result is buggy. And the repair result is worth checking only when the classification returns buggy and the localization result is correct. Our metrics conform to this dependency by ensuring that the performance of the later target is bounded by the previous target.

During model comparison, we first compare precision and recall without manual tuning of the thresholds to purely assess learnability. We prefer higher precision when the recall is comparable. This is because high precision reduces the burden of manual inspection to rule out false positives. When it is necessary to compare the full bug detection ability, we plot precision-recall curves by varying classification thresholds and compare average precision (AP).

\begin{table*}[!t]
  \centering
  \begin{minipage}[t]{\columnwidth}
    \small
\centering
\def\arraystretch{1.2}
\setlength\tabcolsep{6pt}
\caption{Applying our two-phase training on \gnn{} and \buglab{} \cite{pybuglab} (\varmisuse{}).}
\vspace{-2mm}
\label{table:other-varmisuse}
\resizebox{\columnwidth}{!}{\begin{tabular}{@{}lrlrlrlr@{}}
  \toprule
  \multirow{3}{*}{Model} & \multirow{3}{*}{Training Phases} & \multicolumn{2}{c}{\texttt{cls}} & \multicolumn{2}{c}{\texttt{cls-loc}} & \multicolumn{2}{c}{\hspace{2mm}\texttt{cls-loc-rep}} \\
  \cmidrule(lr){3-4} \cmidrule(lr){5-6} \cmidrule(l){7-8}
  & & P & R & P & R & P & R \\
  \midrule
  \gnn{} & Only Synthetic & 1.31 & 36.31 & 0.57 & 15.77 & 0.41 & 11.31 \\
  \gnn{} & Synthetic + Real & 57.14 & 1.19 & 57.14 & 1.19 & 42.86 & 0.89 \\
  \gnn{} & Only \buglab{} & 1.16 & 50.60 & 0.35 & 15.48 & 0.22 & 9.82 \\
  \gnn{} & \buglab{} + Real & \textbf{66.67} & 0.60 & \textbf{66.67} & 0.60 & \textbf{66.67} & 0.60 \\
  \midrule
  Our Model & Synthetic + Real & 64.79 & 13.69 & 61.97 & 13.10 & 56.34 & 11.90 \\
  \bottomrule
\end{tabular}}
  \end{minipage}\hfill
  \begin{minipage}[t]{\columnwidth}
    \small
\centering
\def\arraystretch{1.2}
\setlength\tabcolsep{6pt}
\caption{Applying our two-phase training on \gnn{} and \buglab{} \cite{pybuglab} (\wrongbinop{}).}
\vspace{-2mm}
\label{table:other-wrongbinop}
\resizebox{\columnwidth}{!}{\begin{tabular}{@{}lrlrlrlr@{}}
  \toprule
  \multirow{3}{*}{Model} & \multirow{3}{*}{Training Phases} & \multicolumn{2}{c}{\texttt{cls}} & \multicolumn{2}{c}{\texttt{cls-loc}} & \multicolumn{2}{c}{\hspace{2mm}\texttt{cls-loc-rep}} \\
  \cmidrule(lr){3-4} \cmidrule(lr){5-6} \cmidrule(l){7-8}
  & & P & R & P & R & P & R \\
  \midrule
  \gnn{} & Only Synthetic & 5.59 & 42.57 & 4.79 & 36.46 & 3.64 & 27.70 \\
  \gnn{} & Synthetic + Real & 44.62 & 11.81 & 43.85 & 11.61 & 43.85 & 11.61 \\
  \gnn{} & Only \buglab{} & 3.10 & 55.60 & 2.08 & 37.27 & 1.47 & 26.48 \\
  \gnn{} & \buglab{} + Real & 51.80 & 32.18 & \textbf{51.15} & 31.77 & \textbf{50.82} & 31.57 \\
  \midrule
  Our Model & Synthetic + Real & \textbf{52.30} & 43.99 & 51.09 & 42.97 & 49.64 & 41.75 \\
  \bottomrule
\end{tabular}}
  \end{minipage}
\end{table*}
\begin{figure*}[!t]
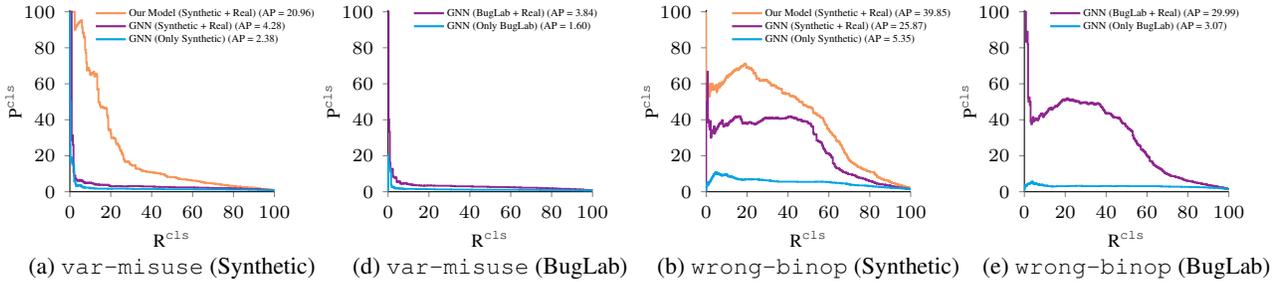

  \centering
  \begin{tikzpicture}
    \begin{groupplot}[
      width=4.3cm, height=4cm,
      compat=newest,
      group style={group size=4 by 1, horizontal sep=43pt},
      tick align=outside,
      major tick length=2pt,
      title style={at={(0.5,-0.25)}, anchor=north, yshift=-10, font=\footnotesize},
      ticklabel style={font=\scriptsize},
      xtick = {0, 20, 40, 60, 80, 100},
      axis x line*=bottom,
      xmin=0.0,
      xmax=100,
      xlabel = {\scriptsize R\textsuperscript{\texttt{cls}}},
      xlabel style = {at={(0.5, -0.16)}},
      ylabel = {\scriptsize P\textsuperscript{\texttt{cls}}},
      ylabel style = {at={(-0.19, 0.5)}},
      ymin=0,
      ymax=101,
      ytick = {0, 20, 40, 60, 80, 100},
      axis y line*=left,
      legend style={
        nodes={scale=0.4, transform shape},
        at={(0.08, 1.05)},
        draw=none,
        anchor=north west,
      },
      legend cell align={left},
      legend image code/.code={
        \draw[line width=0.8pt] plot coordinates {
          (0cm, 0cm)
          (0.2cm, 0cm)
          (0.4cm, 0cm)
        }; 
      }, 
    ]
    \input{figures/rp/gnn/tex/gnn-synthetic-var-misuse}
    \input{figures/rp/gnn/tex/gnn-pybuglab-var-misuse}
    \input{figures/rp/gnn/tex/gnn-synthetic-wrong-binary-operator}
    \input{figures/rp/gnn/tex/gnn-pybuglab-wrong-binary-operator}
    \end{groupplot}
  \end{tikzpicture}
  {\phantomsubcaption\label{fig:pr-gnn-syn-varmisuse}}
  {\phantomsubcaption\label{fig:pr-gnn-bl-varmisuse}}
  {\phantomsubcaption\label{fig:pr-gnn-syn-wrongbinop}}
  {\phantomsubcaption\label{fig:pr-gnn-bl-wrongbinop}}
  \vspace{-5mm}
  \caption{Precision-recall curves and AP for \gnn{} and Our Model with different training phases.}
  \label{fig:pr-gnn}
\end{figure*}
\subsection{Evaluation Results on Two-phase Training}
\label{sec:eval-two-phase}
We present the evaluation results on our two-phase training.

\paragraph{Changing Training Phases}
We create four baselines by only changing the training phases: (i) Only Synthetic: training only on \strain; (ii) Only Real: training only on \rtrain{}; (iii) Mix: combining \strain{} and \rtrain{} into a single training phase; (iv) Two Synthetic: two-phase training, first on \strain{} and then on a new training set constructed by replacing the real bugs in \rtrain{} with synthetic ones while maintaining the imbalance. They are compared with Our Full Method in \cref{table:two-phase-varmisuse,table:two-phase-wrongbinop}. We make the following observations:
\begin{itemize}[leftmargin=*]
  \item Unable to capture data imbalance, Only Synthetic is extremely imprecise (\ie, \textless10\% precision), matching the results from previous works~\cite{pybuglab,DBLP:conf/pldi/HeLRV21}. Such imprecise detectors will flood users with false positives and are practically useless.
  \item For \varmisuse{}, Only Real classifies all test samples as non-buggy. For \wrongbinop{}, Only Real has significantly lower recall than Our Full Method. These results show that our first training phase can make the learning stable or improve model performance.
  \item Mix does not perform well even if it is trained on real bugs. This is because, in the mixed dataset, synthetic bugs outnumber real bugs. As a result, the model does not receive enough learning signals from real bugs.
  \item By capturing data imbalance, Two Synthetic is more precise than Only Synthetic but sacrifices recall. Moreover, Our Full Method reaches significantly higher precision and recall than Two Synthetic, showing the importance of real bugs in training.
\end{itemize}

A precision-recall trade-off exists between Only Synthetic, Mix, and Our Full Method. To fully compare their bug classification capability, we plot their precision-recall curves and AP in \cref{fig:pr-ours}. The results show that Our Full Method significantly outperforms Only Synthetic and Mix with 3-4x higher AP. This means that our two-phase training helps the model generalize better to the real bug distribution.

\paragraph{Varying Data Imbalance Ratio}
To show how the data imbalance in the second training phase helps the model training, we vary the number of non-buggy training samples and keep the buggy training samples the same, resulting in different non-buggy/buggy ratios (2\textsuperscript{0-6} and the original ratio). The results are plotted in \cref{fig:skew}, showing that the non-buggy/buggy ratio affects the precision-recall trade-off. Moreover, AP increases with data imbalance ratio. From 1:1 to the original ratio, AP increases from 7.40 to 20.96 for \varmisuse{} and from 19.84 to 39.85 for \wrongbinop{}.

\begin{table*}[!t]
  \centering
  \begin{minipage}[t]{\columnwidth}
    \small
\centering
\def\arraystretch{1.2}
\setlength\tabcolsep{7pt}
\caption{Evaluating other techniques (\varmisuse{}).}
\vspace{-2mm}
\label{table:others-varmisuse}
\resizebox{\columnwidth}{!}{\begin{tabular}{@{}llrlrlr@{}}
  \toprule
  \multirow{3}{*}{Method} & \multicolumn{2}{c}{\texttt{cls}} & \multicolumn{2}{c}{\texttt{cls-loc}} & \multicolumn{2}{c}{\hspace{2mm}\texttt{cls-loc-rep}} \\
  \cmidrule(lr){2-3} \cmidrule(lr){4-5} \cmidrule(l){6-7}
  & P & R & P & R & P & R \\
  \midrule
  No {\footnotesize \texttt{cls}} Head & 49.45 & 13.39 & 48.35 & 13.10 & 46.15 & 12.50 \\
  No Hierarchy & 51.82 & 16.96 & 50.91 & 16.67 & 48.18 & 15.77 \\
  No Focal Loss & 64.18 & 12.80 & 61.19 & 12.20 & \textbf{58.21} & 11.61 \\
  No Contrastive & 60.00 & 14.29 & 57.50 & 13.69 & 53.75 & 12.80 \\
  \midrule
  Our Full Method & \textbf{64.79} & 13.69 & \textbf{61.97} & 13.10 & 56.34 & 11.90 \\
  \bottomrule
\end{tabular}}
  \end{minipage}\hfill
  \begin{minipage}[t]{\columnwidth}
    \small
\centering
\def\arraystretch{1.2}
\setlength\tabcolsep{7pt}
\caption{Evaluating other techniques (\wrongbinop{}).}
\vspace{-2mm}
\label{table:others-wrongbinop}
\resizebox{\columnwidth}{!}{\begin{tabular}{@{}llrlrlr@{}}
  \toprule
  \multirow{3}{*}{Method} & \multicolumn{2}{c}{\texttt{cls}} & \multicolumn{2}{c}{\texttt{cls-loc}} & \multicolumn{2}{c}{\hspace{2mm}\texttt{cls-loc-rep}} \\
  \cmidrule(lr){2-3} \cmidrule(lr){4-5} \cmidrule(l){6-7}
  & P & R & P & R & P & R \\
  \midrule
  No {\footnotesize \texttt{cls}} Head & 48.20 & 43.58 & 47.30 & 42.77 & 45.95 & 41.55 \\
  No Hierarchy & 48.70 & 45.62 & 47.83 & 44.81 & 46.52 & 43.58 \\
  No Focal Loss & 49.32 & 44.60 & 48.42 & 43.79 & 47.97 & 43.38 \\
  No Contrastive & 47.96 & 43.18 & 47.06 & 42.36 & 46.15 & 41.55 \\
  \midrule
  Our Full Method & \textbf{52.30} & 43.99 & \textbf{51.09} & 42.97 & \textbf{49.64} & 41.75 \\
  \bottomrule
\end{tabular}}
  \end{minipage}
\end{table*}
\paragraph{Varying Amount of Training Data}
We also vary the size of our training sets \strain{} and \rtrain{}. In each experiment, we subsample one training set (with percentage 0, 2, 4, 8, 16, 32, and 64) and fully use the other training set. The subsampling is done by repositories. The results are plotted in \cref{fig:percent}. A general observation is that more training data, in either \strain{} or \rtrain{}, improves AP. More data in \strain{} increases R\textsuperscript{\texttt{cls}}. For \varmisuse{}, the model starts to classify samples as buggy only when given a sufficient amount of data from \strain{}. For \wrongbinop{}, the amount of data in \strain{} does not affect the precision. For \rtrain{}, we can make consistent observations across \varmisuse{} and \wrongbinop{}: first, more data in \rtrain{} improves P\textsuperscript{\texttt{cls}}; second, as the amount of data in \rtrain{} increases from 0, the R\textsuperscript{\texttt{cls}} first decreases but then starts increasing from around 16\%-32\%.

\paragraph{Applying Two-phase Training to Existing Methods}
Next, we demonstrate that our two-phase training method can benefit other methods. We consider \buglab{}, a learned bug selector for injecting synthetic bugs, and its \gnn{} implementation~\cite{pybuglab}. We train four \gnn{} models with different training phases:
\begin{itemize}[leftmargin=*]
  \item Only Synthetic: train only on \strain{}.
  \item Synthetic + Real: two-phase training same as ours, first on \strain{} and then on \rtrain{}.
  \item Only \buglab{}: train only on a balanced dataset where bugs are created by \buglab{}.
  \item \buglab{} + Real: two-phase training, first on a balanced dataset created by \buglab{} and then on \rtrain{}.
\end{itemize}

In \cref{table:other-varmisuse,table:other-wrongbinop}, we show the results of the trained variants together with Our Model (Synthetic + Real) which corresponds to Our Full Method. Comparing models trained with Synthetic + Real, we can see that Our Model clearly outperforms \gnn{} with significantly higher precision and recall. This is likely because Our Model starts from a \cubert{} model pretrained on a large corpus of code. Moreover, compared with Only Synthetic, \gnn{} trained with Synthetic + Real achieves significantly higher precision. The same phenomenon also applies when the first training phase is done with \buglab{}. We provide precision-recall curves and AP of the trained variants in \cref{fig:pr-gnn}, showing that our second training phase can help \gnn{}, trained with either Only Synthetic or Only \buglab{}, achieve higher AP, especially for \wrongbinop{} bugs.

\subsection{Evaluation Results on Other Techniques}
\label{sec:eval-others}
We show the effectiveness of our other techniques with four baselines listed as follows. Each baseline excludes one technique as described below and keeps the other techniques the same as Our Full Method:
\begin{itemize}[leftmargin=*]
  \item No \mytexttt{cls} Head: no classification head. Classification and localization are done jointly like existing pointer models.
  \item No Hierarchy: no task hierarchy. All tasks are performed after the last feature transformation layer.
  \item No Focal Loss: use cross entropy loss for classification.
  \item No Contrastive: no contrastive learning with $\beta = 0$.
\end{itemize}
The above baselines have similar-level recall but noticeably lower precision than Our Full Method. This means all the evaluated techniques contribute to the high performance of Our Full Method. The classification head and task hierarchy play a major role for \varmisuse{}. We provide results with different task orders and $\beta$ values in \cref{sec:more-eval}.

\subsection{Scanning Latest Open Source Repositories}
\label{sec:scan}
In an even more practical setting, we evaluate our method on the task of scanning the latest version of open source repositories. To achieve this, we obtain 1118 (\resp, 2339) GitHub repositories for \varmisuse{} and \wrongbinop{} (\resp, \argswap{}). Those repositories do not overlap with the ones used to construct \strain{} and \rtrain{}. We apply our full method on all eligible functions in the repositories, without any sample filtering or threshold tuning, and deduplicate the reported warnings together with the extracted real bugs. This results in 427 warnings for \varmisuse{}, 2102 for \wrongbinop{}, and 203 for \argswap{}.

\begin{table}
  \small
  \centering
  \def\arraystretch{1.1}
  \setlength\tabcolsep{7pt}
  \caption{Manual inspection result on the reported warnings.}
  \vspace{-2mm}
  \label{table:inspection}
  \resizebox{\columnwidth}{!}{\begin{tabular}{@{}lrrr@{}}
    \toprule
    Bug Type & Bugs & Quality Issues & False Positives \\
    \midrule
    \varmisuse{} & 50 & 10 & 40 \\
    \texttt{\footnotesize wrong-binop} & 6 & 80 & 14 \\
    \texttt{\footnotesize wrong-binop-filter} & 37 & 11 & 52 \\
    \argswap{} & 17 & 3 & 80 \\
    \bottomrule
  \end{tabular}}
\end{table}
For each bug type, we manually investigate $100$ randomly sampled warnings and, following \cite{DBLP:journals/pacmpl/PradelS18}, categorize them into (i) \emph{Bugs}: warnings that cause wrong program behaviors, errors, or crashes; (ii) \emph{Code Quality Issues}: warnings that are not bugs but impair code quality (\eg, unused variables), or do not conform to Python coding conventions, and therefore should be raised and fixed; (iii) \emph{False Positives}: the rest. To reduce human bias, two authors independently assessed the warnings and discussed differing opinions to reach an agreement. We show the inspection statistics in \cref{table:inspection} and present case studies in \cref{sec:cases}. Moreover, we report a number of bugs to the developers and the links to the bug reports are provided in \cref{sec:reports}.


For \varmisuse{}, most code quality issues are unused variables. For \wrongbinop{}, most warnings are related to \code{==}, \code{!=}, \code{is}, or \code{is not}. Our detector flags the use of \code{==} and \code{!=} for comparing with \code{None}, and the use of \code{is} (\resp, \code{is not}) for equality (\resp, inequality) check with primitive types. Those behaviors, categorized by us as code quality issues, do not conform to Python coding conventions and even cause bugs in rare cases~\cite{sof}. Our model learns to detect them because such samples exist as real bugs in \rtrain{}. Depending on the demand on code quality, users might want to turn off such warnings. We simulate this case by filtering out those behaviors and inspect another 100 random warnings from the 255 warnings after filtering. The results are shown in row \mytexttt{wrong-binop-filter} of \cref{table:inspection}. The bug ratio becomes significantly higher than the original version. For \argswap{}, our model mostly detects bugs with Python standard library functions, such as \mytexttt{isinstance} and \mytexttt{super}, or APIs of popular libraries such as TensorFlow. Most false positives are reported on repository-specific functions not seen during training.

The inspection results demonstrate that our detectors are performant and useful in practice. Counting both bugs and code quality issues as true positives, the precision either matches the evaluation results with \rtest{} (\varmisuse{} and \wrongbinop{}) or the performance gap discussed in \cref{sec:intro} is greatly reduced (\argswap{}). This demonstrates that our method is able to handle the real bug distribution and our dataset can be reliably used for measuring the practical effectiveness of bug detectors.

\section{Related Work}
We now discuss works most closely related to ours.

\paragraph{Machine Learning for Bug Detection}
\gnn{}s~\cite{DBLP:conf/iclr/AllamanisBK18}, LSTMs~\cite{DBLP:conf/iclr/VasicKMBS19}, and \great{}~\cite{DBLP:conf/iclr/HellendoornSSMB20} are used to detect \varmisuse{} bugs. Deepbugs~\cite{DBLP:journals/pacmpl/PradelS18} learns classifiers based on code embeddings to detect \wrongbinop{}, \argswap{}, and incorrect operands bugs. Hoppity~\cite{DBLP:conf/iclr/DinellaDLNSW20} learns to perform graph transformations representing small code edits. A number of models are proposed to handle multiple coding tasks including bug detection. This includes PLUR~\cite{plur}, a unified graph-based framework for code understanding, and pre-trained code models such as \cubert{}~\cite{DBLP:conf/icml/KanadeMBS20} and CodeBert~\cite{DBLP:conf/emnlp/FengGTDFGS0LJZ20}.

The above works mainly use datasets with synthetic bugs to train and evaluate the learned detectors. Some spend efforts on evaluation with real bugs but none of them completely capture the real bug distribution: the authors of~\cite{DBLP:conf/iclr/VasicKMBS19} and~\cite{DBLP:conf/iclr/HellendoornSSMB20} evaluate their models on a small set of paired code changes from GitHub (\ie, buggy/non-buggy ratio is $1$:$1$). The PyPIBugs~\cite{pybuglab} and ManySStuBs4J~\cite{DBLP:conf/msr/KarampatsisS20} datasets use real bugs from GitHub commits but do not contain non-buggy samples. Hoppity~\cite{DBLP:conf/iclr/DinellaDLNSW20} is trained and evaluated on small code edits in GitHub commits, which are not necessarily bugs and can be refactoring, version changes, and other code changes~\cite{DBLP:conf/icml/BerabiHRV21}. Compared with the above datasets, our datasets with real bugs are the closest to the real bug distribution so far.

Other works focus on complex bugs such as security vulnerabilities \cite{DBLP:conf/ndss/LiZXO0WDZ18,DBLP:conf/nips/ZhouLSD019,9699412}. We believe that the characteristics of bugs discussed in our work are general and extensible to complex bugs.

\paragraph{Distribution Shift in Bug Detection and Repair}
A few works try to create realistic bugs for training bug detectors or fixers. \buglab{}~\cite{pybuglab} jointly learns a bug selector with the detector to create bugs for training. Since no real bugs are involved in the training process, it is unclear if the learned selector actually constructs realistic bugs. Based on code embeddings, SemSeed~\cite{DBLP:conf/sigsoft/PatraP21} learns manually defined bug patterns from real bugs, to create new, realistic bugs, which can be used to train bug detectors. Unlike SemSeed, our bug detectors learn directly from real bugs which avoids one level of information loss. BIFI~\cite{DBLP:conf/icml/YasunagaL21} jointly learns a breaker for injecting errors to code and a fixer for fixing errors. Focusing on fixing parsing and compilation errors, BIFI assumes a perfect external error classifier (\eg, AST parsers and compilers), while our work learns a classifier for software bugs. Namer~\cite{DBLP:conf/pldi/HeLRV21} proposes a similar two-step learning recipe for finding naming issues. Different from our work, Namer relies on manually defined patterns and does not benefit from training with synthetic bugs.

\paragraph{Neural Models of Code}
Apart from bug detection, neural models are adopted for a number of other code tasks including method name suggestion~\cite{DBLP:journals/pacmpl/AlonZLY19,DBLP:conf/icml/AllamanisPS16,DBLP:conf/iclr/ZugnerKCLG21}, type inference~\cite{DBLP:conf/iclr/WeiGDD20,DBLP:conf/pldi/AllamanisBDG20}, code editing~\cite{DBLP:journals/pacmpl/Brody0Y20,DBLP:conf/iclr/YinNABG19}, and program synthesis~\cite{DBLP:conf/icml/0002SLY20,DBLP:conf/iclr/BrockschmidtAGP19,DBLP:journals/corr/abs-2111-01633}. More recently, large language models are used to generate real-world code~\cite{DBLP:journals/corr/abs-2108-07732,DBLP:journals/corr/abs-2107-03374}. 

\paragraph{Code Rewriting for Data Augmentation}
Semantics-preserving code rewriting is used for producing programs, \eg{}, for adversarial training of type inference models \cite{DBLP:conf/icml/BielikV20} and contrastive learning of code clone detectors \cite{DBLP:conf/emnlp/0001JZA0S21}. The rewritten and the original programs are considered to be similar. In the setting of bug detection, however, the programs created by bug-injection rules and the original ones should be considered by the model to be distinct \cite{DBLP:conf/sigsoft/PatraP21,pybuglab}, which is captured by our contrastive loss.

\section{Conclusion}

In this work, we revealed a fundamental mismatch between the real bug distribution and the synthetic bug distribution used to train and evaluate existing learning-based bug detectors. To mitigate this distribution shift, we proposed a two-phase learning method combined with a task hierarchy, focal loss, and contrastive learning. Our evaluation demonstrates that the method yields bug detectors able to capture the real bug distribution. We believe that our work is an important step towards understanding the complex nature of bug detection and learning practically useful bug detectors.


\bibliography{bib}
\bibliographystyle{icml2022}

\newpage
\appendix
\onecolumn

\section{Computing Localization and Repair Probabilities}
\label{sec:compute-repair}

Now we discuss how pointer models compute the localization probabilities $P^{loc} = [p^{loc}_1, \ldots, p^{loc}_n]$ and the repair probabilities $P^{rep} = [p^{rep}_1, \ldots, p^{rep}_l]$. Given the feature embeddings $[h_1, \ldots, h_n]$ for program tokens $T=\langle t_1, t_2, \ldots, t_n \rangle$, pointer models first compute a score vector $S^{loc} = [s^{loc}_1, \ldots, s^{loc}_n]$ where each score $s^{loc}_i$ reflects the likelihood of token $t_i$ to be the bug location. If $t_i \in Loc$, \ie, token $t_i$ is a candidate bug location, a feedforward network $\pi^{loc}: \mathbb{R}^m \rightarrow \mathbb{R}$ is applied on the feature vector $h_i$ to compute $s^{loc}_i$. Otherwise, it is unlikely that $t_i$ is the bug location so a minus infinity score is assigned. Formally, $s^{loc}_i$ is computed as follows:
\begin{equation*}
  s^{loc}_i =
  \begin{cases} 
    \pi^{loc}(h_i) & \text{if } M^{loc}[i] = 1, \\
    -\text{inf} & \text{otherwise},
  \end{cases}
\end{equation*}
where $M^{loc}$ is the localization candidate mask:
\begin{equation*}
  M^{loc}[i] =
  \begin{cases}
    1 & \text{if } t_i \in Loc, \\
    0 & \text{otherwise}.
  \end{cases}
\end{equation*}
$S^{loc}$ is then normalized to localization probabilities with the softmax function: $P^{loc} = [p^{loc}_1, \ldots, p^{loc}_n] = \softmax(S^{loc})$.

Depending on the bug type, the set of repair tokens, $Rep$, can be drawn from $T$ (\eg, for \varmisuse{} and \argswap{}) or fixed (\eg, for \wrongbinop). For the former case, $P^{rep}$ is computed in the same way as computing $P^{loc}$, except that another feedforward network, $\pi^{rep}$, and the repair candidate mask, $M^{rep}$, are used instead of $\pi^{loc}$ and $M^{loc}$. When $Rep$ is a fixed set of $l$ tokens, the repair prediction is basically an $l$-class classification problem. We treat the first token $t_1$ of $T$ as the repair token $t_{[rep]}$ and apply $\pi^{rep}: \mathbb{R}^m \rightarrow \mathbb{R}^l$ over its feature vector $h_{[rep]}$ to compute scores $S = \pi^{rep}(h_{[rep]})$. Then, the final repair score is set to $S[i]$ if $M^{rep}$ indicates that the $i$-th repair token is valid, or to minus infinity otherwise. Overall, the repair scores $S^{rep} = [s^{rep}_1, \ldots, s^{rep}_l]$ are computed as follows:
\begin{equation*}
  s^{rep}_i =
  \begin{cases} 
    S[i] & \text{if } M^{rep}[i] = 1, \\
    -\text{inf} & \text{otherwise},
  \end{cases}
\end{equation*}
$S^{rep}$ is then normalized to repair probabilities with the softmax function: $P^{rep} = [p^{rep}_1, \ldots, p^{loc}_l] = \softmax(S^{rep})$.

\section{Implementation, Model and Training Details}
\label{sec:details}
In this section, we provide details in implementation, models, and training.

\paragraph{Constructing the Test Sets in \cref{figure:shift}}
In \cref{figure:shift}, we mention three test sets used to reveal distribution shift in existing learning-based \varmisuse{} detectors. Test set I is a balanced dataset with synthetic bugs, created by randomly selecting 336 non-buggy samples from \rtest{} and injecting one synthetic bug into each non-buggy sample. Test set II is a balanced dataset with real bugs, created by replacing the synthetic bugs in the first test set by the 336 real bugs in \rtest{}. Test set III is \rtest{}.

\paragraph{Three Bug Types Handled by Our Work}
The definition of \varmisuse{} (\resp, \wrongbinop{} and \argswap{}) can be found in~\cite{DBLP:conf/iclr/AllamanisBK18,DBLP:conf/iclr/VasicKMBS19} (\resp, in \cite{DBLP:journals/pacmpl/PradelS18}). To determine $Loc$ and $Rep$, we mainly follow~\cite{pybuglab,DBLP:conf/icml/KanadeMBS20} and add small adjustments to capture more real bugs:
\begin{itemize}[leftmargin=*]
  \item \varmisuse{}: we include all appearances of all local variables in $Loc$, as long as the appearance is not in a function definition and the variable has been defined before the appearance. When constructing $Rep$ for each bug location variable, we include all local variable definitions that can be found in the scope of the bug location variable, except for the ones that define the bug location variable itself.
  \item \wrongbinop{}: we deal with three sets of binary operators: arithmetics \{\code{+}, \code{*}, \code{-}, \code{/}, \code{\%}\}, comparisons \{\code{==}, \code{!=}, \code{is}, \code{is not}, \code{\textless}, \code{\textless=}, \code{\textgreater}, \code{\textgreater=}, \code{in}, \code{not in}\}, and booleans \{\code{and}, \code{or}\}. If a binary operator belongs to any of the three sets, it is added to $Loc$. The set that the operator belongs to, excluding the operator itself, is treated as $Rep$. The repair candidate mask $M^{rep}$ is of size 17, \ie, it includes all the operators in the three sets. $M^{rep}$ sets the operators in $Rep$ to $1$ and the other operators to $0$.
  \item \argswap{}: We handle most function arguments but exclude keyworded and variable-length arguments that are less likely to be mistaken. In contrast to the other bug types, we also support the swapping of arguments that consist of more than a single token (\eg., an expression), by simply marking the first token as the bug location or the repair token. Moreover, we consider only functions that have two or more handled arguments. We put all candidate arguments in $Loc$. For each argument in $Loc$, $Rep$ is the other candidate arguments used in the same function.
\end{itemize}
For bug injection and real bug extraction, we apply the bug-inducing rewriting rules in~\cite{pybuglab} given the definitions of $Loc$ and $Rep$ above.

\paragraph{Implementation with \cubert{}}
Here, we describe the implementation of our techniques with \cubert{}. \cubert{} tokenizes the input program into a sequence of sub-tokens. When constructing the masks $M^{\text{loc}}$, $C^{\text{loc}}$, $M^{\text{rep}}$, and $C^{\text{rep}}$ from $Loc$ and $Rep$, we set the first sub-token of each token to $1$. As standard with BERT-like models~\cite{DBLP:conf/naacl/DevlinCLT19}, the first sub-token of the input sequence to \cubert{} is always \mytexttt{[CLS]} used as aggregate sequence representation for classification tasks. We also use this token and its corresponding feature embedding for bug classification (all three tasks) and repair (only \wrongbinop{}). \cubert{} consists of a sequence of BERT layers and thus naturally aligns with our task hierarchy. Our two-phase training is technically a two-phase fine-tuning procedure when applied to pre-trained models like \cubert{}. \cubert{} requires the input sequence to be of fixed length, meaning that shorted sequences will be padded and longer sequences will be truncated. We chose length 512 due to contraints on hardware: \cubert{} is demanding in terms of GPU memory and longer lengths caused out-of-memory errors on our machines. When extracting real bugs and injecting bugs into open source programs, we only consider bugs for which the bug location and at least one correct repair token are within the fixed length. This includes most real bugs we found.

\paragraph{Model Details}
\cubert{} is a BERT-Large model with 24 hidden layers, 16 attention heads, 1024 hidden units, and in total ~340M parameters. Our classification head $\pi^{cls}$ is a two-layer feedforward network. The localization head $\pi^{loc}$ is just a linear layer. The repair head $\pi^{rep}$ is a linear layer for \varmisuse{} and \argswap{} and a two-layer feedforward network for \wrongbinop{}. The size of the hidden layers is 1024 for all task heads. The implementation of our model is based on Hugging Face~\cite{DBLP:journals/corr/abs-1910-03771} and PyTorch~\cite{DBLP:conf/nips/PaszkeGMLBCKLGA19}.

\begin{table}[!t]
  \small
  \centering
  \def\arraystretch{1.2}
  \setlength\tabcolsep{10pt}
  \caption{The number of epochs, learning rate (LR), and time cost for the two training phases.}
  \vspace{-2mm}
  \label{table:training}
  \begin{tabular}{@{}llrrrrr@{}}
    \toprule
    \multirow{3}{*}{Bug Type} & \multicolumn{3}{c}{First phase} & \multicolumn{3}{c}{Second phase} \\
    \cmidrule(lr){2-4} \cmidrule(l){5-7}
    & Epochs & LR & Time & Epochs & LR & Time \\
    \midrule
    \varmisuse{} & 1 & 10\textsuperscript{-6} & 15h & 2 & 10\textsuperscript{-6} & 10h \\
    \wrongbinop{} & 1 & 10\textsuperscript{-5} & 15h & 2 & 10\textsuperscript{-6} & 6h \\
    \argswap{} & 1 & 10\textsuperscript{-5} & 15h & 1 & 10\textsuperscript{-6} & 2h \\
    \bottomrule
  \end{tabular}
\end{table}
\paragraph{Training Details}
Our experiments were done on servers with NVIDIA RTX 2080 Ti and NVIDIA TITAN X GPUs. As described in \cref{sec:two-step}, our training procedure consists of two phases. In the first phase, we load a pretrained \cubert{} model provided by the authors~\cite{DBLP:conf/icml/KanadeMBS20} and fine-tune it with \strain{}. In the second phase, we load the model trained from the first phase and perform fresh fine-tuning with \rtrain{}. The number of epochs, learning rate, and the time cost of the two training phases are shown in \cref{table:training}. Both training phases require at most two epochs to achieve good performance, highlighting the power of pretrained models to quickly adapt to new tasks and data distributions. In each batch, we feed two samples into the model as larger batch size will cause out-of-memory errors.

For fair comparison, when creating synthetic bugs with \buglab{}, we do not perform their data augmentation rewrite rules for all models. Those rules apply to all models and can be equally beneficial. When training \gnn{} models with \strain{} and \rtrain{}, we follow~\cite{pybuglab} to use early stopping over \rval{}. When training with \buglab{}, we use 80 meta-epochs, 5k samples (buggy/non-buggy raito 1:1) per meta-epoch, and 40 model training epochs within one meta-epoch. This amounts to a total of around $6$ days of training time for \gnn{}.

\section{More Evaluation Results}
\label{sec:more-eval}
In this section, we present additional evaluation results.

\paragraph{Evaluation Results for \argswap{}}
We repeat the experiments in \cref{sec:eval-two-phase,sec:eval-others} for \argswap{}. The results are shown in \cref{table:two-phase-argswap,table:related-argswap,table:others-argswap,fig:pr-argswap,fig:skew-argswap,fig:percent-argswap,fig:pr-related-argswap}. Most observations that we can make from those results are similar to what we discussed in \cref{sec:eval-two-phase,sec:eval-others} for \varmisuse{} and \wrongbinop{}. We highlight two difference: first, Our Full Method does not have a clear advantage over Only Synthetic and Mix in terms of AP (see \cref{fig:pr-argswap}); second, the data imbalance and the amount of training data do not clearly improve the AP (see \cref{fig:skew-argswap,fig:percent-argswap}). These different points are likely due to the distinct characteristics of \argswap{} bugs. We leave it as an interesting future work item to further improve the performance of \argswap{} detectors.

\paragraph{Parameter Selection for Task Hierarchy and Contrastive Learning}
In \cref{table:contrastive-varmisuse,table:contrastive-wrongbinop,table:contrastive-argswap} (\resp, \cref{table:hierarchy-varmisuse,table:hierarchy-wrongbinop,table:hierarchy-argswap}), we show the model performance by only changing weight $\beta$ of the contrastive loss (\resp, the task order in our task hierarchy). For \varmisuse{} and \wrongbinop{}, Our Full Method (highlighted with $\star$) performs the best among all the configurations. In terms of $\beta$ and \varmisuse{}, Our Full Method ($\beta=0.5$) is less precise but has significantly higher recall than $\beta=8$. For \argswap{}, Our Full Method performs the best on the validation set but not on the test set.

\clearpage
\begin{table*}[!t]
  \centering
  \begin{minipage}[t]{0.48\columnwidth}
    \centering
    \begin{minipage}[t]{\textwidth}
      \small
\centering
\def\arraystretch{1.2}
\setlength\tabcolsep{7pt}
\caption{Changing training phases (\argswap{}).}
\vspace{-2mm}
\label{table:two-phase-argswap}
\resizebox{\columnwidth}{!}{\begin{tabular}{@{}llrlrlr@{}}
  \toprule
  \multirow{3}{*}{Method} & \multicolumn{2}{c}{\texttt{cls}} & \multicolumn{2}{c}{\texttt{cls-loc}} & \multicolumn{2}{c}{\hspace{2mm}\texttt{cls-loc-rep}} \\
  \cmidrule(lr){2-3} \cmidrule(lr){4-5} \cmidrule(l){6-7}
  & P & R & P & R & P & R \\
  \midrule
  Only Synthetic & 1.31 & 39.84 & 1.00 & 30.49 & 0.79 & 23.98 \\
  Only Real & 0 & 0 & 0 & 0 & 0 & 0 \\
  Mix & 2.02 & 32.93 & 1.69 & 27.64 & 1.59 & 26.02 \\ 
  Two Synthetic & 44.19 & 7.72 & 44.19 & 7.72 & 44.19 & 7.72 \\
  \midrule
  Our Full Method & \textbf{73.68} & 5.69 & \textbf{73.68} & 5.69 & \textbf{73.68} & 5.69 \\
  \bottomrule
\end{tabular}}
      \vspace{5mm}
    \end{minipage}
    \medskip
    \begin{minipage}[t]{\textwidth}
      \small
\centering
\def\arraystretch{1.2}
\setlength\tabcolsep{6pt}
\caption{Applying our two-phase training on \gnn{} and \buglab{} \cite{pybuglab} (\argswap{}).}
\vspace{-2mm}
\label{table:related-argswap}
\resizebox{\columnwidth}{!}{\begin{tabular}{@{}lrlrlrlr@{}}
  \toprule
  \multirow{3}{*}{Model} & \multirow{3}{*}{Training Phases} & \multicolumn{2}{c}{\texttt{cls}} & \multicolumn{2}{c}{\texttt{cls-loc}} & \multicolumn{2}{c}{\hspace{2mm}\texttt{cls-loc-rep}} \\
  \cmidrule(lr){3-4} \cmidrule(lr){5-6} \cmidrule(l){7-8}
  & & P & R & P & R & P & R \\
  \midrule
  \gnn{} & Only Synthetic & 0.99 & 50.00 & 0.68 & 34.15 & 0.43 & 21.95 \\
  \gnn{} & Synthetic + Real & \textbf{83.33} & 4.07 & \textbf{83.33} & 4.07 & \textbf{75.00} & 3.66 \\
  \gnn{} & Only \buglab{} & 0.81 & 51.63 & 0.50 & 32.11 & 0.37 & 23.58 \\
  \gnn{} & \buglab{} + Real & 81.82 & 3.66 & 81.82 & 3.66 & 72.73 & 3.25 \\
  \midrule
  Our Model & Synthetic + Real & 73.68 & 5.69 & 73.68 & 5.69 & 73.68 & 5.69 \\
  \bottomrule
\end{tabular}} 
      \vspace{3mm}
    \end{minipage}
    \medskip
    \begin{minipage}[t]{\textwidth}
      \small
\centering
\def\arraystretch{1.2}
\setlength\tabcolsep{7pt}
\caption{Evaluating other techniques (\argswap{}).}
\vspace{-2mm}
\label{table:others-argswap}
\resizebox{\columnwidth}{!}{\begin{tabular}{@{}llrlrlr@{}}
  \toprule
  \multirow{3}{*}{Method} & \multicolumn{2}{c}{\texttt{cls}} & \multicolumn{2}{c}{\texttt{cls-loc}} & \multicolumn{2}{c}{\hspace{2mm}\texttt{cls-loc-rep}} \\
  \cmidrule(lr){2-3} \cmidrule(lr){4-5} \cmidrule(l){6-7}
  & P & R & P & R & P & R \\
  \midrule
  No {\footnotesize \texttt{cls}} Head & 34.21 & 5.28 & 34.21 & 5.28 & 34.21 & 5.28  \\
  No Hierarchy & 61.29 & 7.72 & 61.29 & 7.72 & 61.29 & 7.72 \\
  No Focal Loss & \textbf{73.68} & 5.69 & \textbf{73.68} & 5.69 & \textbf{73.68} & 5.69 \\
  No Contrastive & 46.15 & 7.32 & 46.15 & 7.32 & 46.15 & 7.32 \\
  \midrule
  Our Full Method & \textbf{73.68} & 5.69 & \textbf{73.68} & 5.69 & \textbf{73.68} & 5.69 \\
  \bottomrule
\end{tabular}}
    \end{minipage}
  \end{minipage}\hfill
  \begin{minipage}[t]{0.48\columnwidth}
    \centering
    \begin{minipage}[t]{\textwidth}
      \begin{minipage}[t]{0.48\textwidth}
        \centering
\begin{tikzpicture}
  \begin{groupplot}[
    width=4.3cm, height=4cm,
    compat=newest,
    group style={group size=1 by 1, horizontal sep=35pt},
    title style={at={(0.5,-0.25)}, anchor=north, yshift=-10, font=\footnotesize},
    ticklabel style={font=\tiny},
    tick align=outside,
    major tick length=2pt,
    xtick = {0, 20, 40, 60, 80, 100},
    axis x line*=bottom,
    xmin=0.0,
    xmax=100,
    xlabel = {\scriptsize R\textsuperscript{\texttt{cls}}},
    xlabel style = {at={(0.5, -0.15)}},
    ylabel = {\scriptsize P\textsuperscript{\texttt{cls}}},
    ylabel style = {at={(-0.17, 0.5)}},
    ymin=0,
    ymax=101,
    ytick = {0, 20, 40, 60, 80, 100},
    axis y line*=left,
    legend style={
      nodes={scale=0.4, transform shape},
      at={(0.08, 1.05)},
      draw=none,
      anchor=north west,
    },
    legend cell align={left},
    legend image code/.code={
      \draw[line width=0.8pt] plot coordinates {
        (0cm, 0cm)
        (0.2cm, 0cm)
        (0.4cm, 0cm)
      };
    }, 
  ]
  \nextgroupplot
    \input{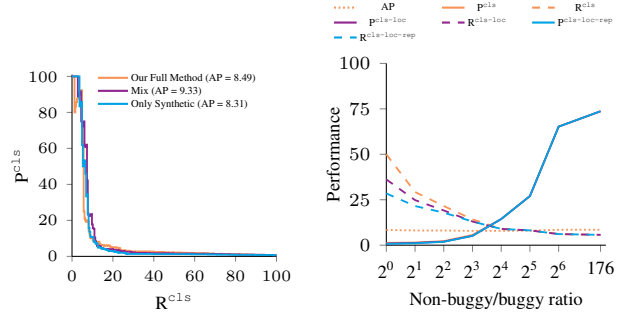}
  \end{groupplot}
\end{tikzpicture}
\vspace{-6mm}
\captionof{figure}{Precision-recall curve and AP for methods in \cref{table:two-phase-argswap} (\argswap{}).}
\label{fig:pr-argswap}
      \end{minipage}
      \hfill
      \begin{minipage}[t]{0.48\textwidth}
        \centering
\begin{tikzpicture}
  \begin{groupplot}[
    width=4.5cm, height=4cm,
    compat=newest,
    group style={group size=1 by 1, horizontal sep=38pt},
    tick align=outside,
    major tick length=2.5pt,
    title style={at={(0.5,-0.31)}, anchor=north, yshift=-10, font=\footnotesize},
    ticklabel style={font=\scriptsize},
    axis x line*=bottom,
    xmin=1,
    xmax=200,
    xmode=log,
    log basis x=2,
    xlabel = {\scriptsize Non-buggy/buggy ratio},
    xlabel style = {at={(0.5, -0.21)}},
    ylabel = {\scriptsize Performance},
    ylabel style = {at={(-0.17, 0.5)}},
    ymin=0,
    ymax=100,
    ytick = {0, 25, 50, 75, 100},
    axis y line*=left,
    legend style={
      nodes={scale=0.4, transform shape},
      at={(0.4,1.38)},
      anchor=north,
      legend columns=3,
      draw=none,
      align=center,
      /tikz/every even column/.append style={column sep=0.3cm},
    },
    legend image code/.code={
      \draw[line width=0.8pt] plot coordinates {
        (0cm, 0cm)
        (0.15cm, 0cm)
        (0.3cm, 0cm)
      };
    }, 
  ]
    \nextgroupplot[
      xtick = {1, 2, 4, 8, 16, 32, 64, 176},
      xticklabels = {2\textsuperscript{0}, 2\textsuperscript{1}, 2\textsuperscript{2}, 2\textsuperscript{3}, 2\textsuperscript{4}, 2\textsuperscript{5}, 2\textsuperscript{6}, 176},
    ]
      \input{figures/skew/argswap}
    \legend{$\text{AP}$, $\text{P}^{\texttt{cls}}$, $\text{R}^{\texttt{cls}}$, $\text{P}^{\texttt{cls-loc}}$, $\text{R}^{\texttt{cls-loc}}$, $\text{P}^{\texttt{cls-loc-rep}}$, $\text{R}^{\texttt{cls-loc-rep}}$}
  \end{groupplot}
\end{tikzpicture}
\vspace{-6mm}
\captionof{figure}{Varying data skewness in the second training phase (\argswap{}).}
\label{fig:skew-argswap}
      \end{minipage}
      \vspace{5mm}
    \end{minipage}
    \medskip
    \begin{minipage}[t]{\textwidth}
      \centering
\begin{tikzpicture}
  \begin{groupplot}[
    width=4.4cm, height=4cm,
    compat=newest,
    group style={group size=2 by 1, horizontal sep=40pt},
    tick align=outside,
    major tick length=2pt,
    title style={at={(0.5,-0.25)}, anchor=north, yshift=-10, font=\footnotesize},
    ticklabel style={font=\scriptsize},
    xtick = {0.8, 2, 4, 8, 16, 32, 64, 100},
    xticklabels = {0, 2\textsuperscript{1}, 2\textsuperscript{2}, 2\textsuperscript{3}, 2\textsuperscript{4}, 2\textsuperscript{5}, 2\textsuperscript{6}, 100},
    axis x line*=bottom,
    xmin=0.8,
    xmax=100,
    xmode=log,
    log basis x=2,
    axis x discontinuity=crunch,
    xlabel = {\scriptsize Percentage},
    xlabel style = {at={(0.5, -0.15)}},
    ylabel = {\scriptsize Performance},
    ylabel style = {at={(-0.17, 0.5)}},
    ymin=0,
    ymax=100,
    ytick = {0, 25, 50, 75, 100},
    axis y line*=left,
    legend style={
      nodes={scale=0.5, transform shape},
      at={(-0.3,1.4)},
      anchor=north,
      legend columns=4,
      draw=none,
      align=center,
      /tikz/every even column/.append style={column sep=0.4cm},
    },
    legend image code/.code={
      \draw[line width=1.5pt] plot coordinates {
        (0cm, 0cm)
        (0.26cm, 0cm)
        (0.52cm, 0cm)
      };
    }, 
  ]
    \nextgroupplot[
      title={(a) \texttt{\footnotesize synthetic-train}},
    ]
      \input{figures/synthetic/argswap}

    \nextgroupplot[
      title={(b) \texttt{\footnotesize real-train}},
    ]
      \input{figures/real/argswap}

    \legend{$\text{AP}$, $\text{P}^{\texttt{cls}}$, $\text{R}^{\texttt{cls}}$, $\text{P}^{\texttt{cls-loc}}$, $\text{R}^{\texttt{cls-loc}}$, $\text{P}^{\texttt{cls-loc-rep}}$, $\text{R}^{\texttt{cls-loc-rep}}$}
  \end{groupplot}
\end{tikzpicture}
{\phantomsubcaption\label{fig:synthetic-argswap}}
{\phantomsubcaption\label{fig:real-argswap}}
\vspace{-6mm}
\captionof{figure}{Model performance with subsampled \strain{} or \rtrain{} (\argswap{}).}
\label{fig:percent-argswap}
      \vspace{3mm}
    \end{minipage}
    \medskip
    \begin{minipage}[t]{\textwidth}
      \centering
\begin{tikzpicture}
  \begin{groupplot}[
    width=4.3cm, height=4cm,
    compat=newest,
    group style={group size=2 by 1, horizontal sep=43pt},
    tick align=outside,
    major tick length=2pt,
    title style={at={(0.5,-0.25)}, anchor=north, yshift=-10, font=\footnotesize},
    ticklabel style={font=\scriptsize},
    xtick = {0, 20, 40, 60, 80, 100},
    axis x line*=bottom,
    xmin=0.0,
    xmax=100,
    xlabel = {\scriptsize R\textsuperscript{\texttt{cls}}},
    xlabel style = {at={(0.5, -0.15)}},
    ylabel = {\scriptsize P\textsuperscript{\texttt{cls}}},
    ylabel style = {at={(-0.17, 0.5)}},
    ymin=0,
    ymax=101,
    ytick = {0, 20, 40, 60, 80, 100},
    axis y line*=left,
    legend style={
      nodes={scale=0.4, transform shape},
      at={(0.08, 1.05)},
      draw=none,
      anchor=north west,
    },
    legend cell align={left},
    legend image code/.code={
      \draw[line width=0.8pt] plot coordinates {
        (0cm, 0cm)
        (0.2cm, 0cm)
        (0.4cm, 0cm)
      }; 
    }, 
  ]
  \input{figures/rp/gnn/tex/gnn-synthetic-argument-swap}
  \input{figures/rp/gnn/tex/gnn-pybuglab-argument-swap}
  \end{groupplot}
\end{tikzpicture}
{\phantomsubcaption\label{fig:pr-gnn-syn-argswap}}
{\phantomsubcaption\label{fig:pr-gnn-bl-argswap}}
\vspace{-6mm}
\captionof{figure}{Precision-recall curves and AP for \gnn{} and Our Model with different training phases (\argswap{}).}
\label{fig:pr-related-argswap}
    \end{minipage}
  \end{minipage}
\end{table*}

\clearpage
\begin{table*}[!t]
  \centering
  \begin{minipage}[b]{0.48\columnwidth}
    \begin{minipage}[t]{\columnwidth}
      \small
\centering
\def\arraystretch{1.2}
\setlength\tabcolsep{7pt}
\caption{Different weight $\beta$ (\varmisuse{}).}
\vspace{-2mm}
\label{table:contrastive-varmisuse}
\resizebox{\columnwidth}{!}{\begin{tabular}{@{}llrlrlr@{}}
  \toprule
  \multirow{2.5}{*}{\shortstack[l]{Weight $\beta$ of\\Contrastive Loss}} & \multicolumn{2}{c}{\texttt{cls}} & \multicolumn{2}{c}{\texttt{cls-loc}} & \multicolumn{2}{c}{\hspace{2mm}\texttt{cls-loc-rep}} \\
  \cmidrule(lr){2-3} \cmidrule(lr){4-5} \cmidrule(l){6-7}
  & P & R & P & R & P & R \\
  \midrule
  0 & 60.00 & 14.29 & 57.50 & 13.69 & 53.75 & 12.80 \\
  0.25 & 59.21 & 13.39 & 56.58 & 12.80 & 55.26 & 12.50 \\
  $\star$ 0.5 & 64.79 & 13.69 & 61.97 & 13.10 & 56.34 & 11.90 \\
  1 & 61.90 & 11.61 & 57.14 & 10.71 & 55.56 & 10.42 \\
  2 & 61.67 & 11.01 & 58.33 & 10.42 & 58.33 & 10.42 \\
  4 & 58.62 & 10.12 & 55.17 & 9.52 & 51.72 & 8.93 \\
  8 & \textbf{71.43} & 1.49 & \textbf{71.43} & 1.49 & \textbf{71.43} & 1.49 \\
  16 & 63.64 & 6.25 & 63.64 & 6.25 & 60.61 & 5.95 \\
  \bottomrule
\end{tabular}}
      \vspace{3mm}
    \end{minipage}
    \medskip
    \begin{minipage}[t]{\columnwidth}
      \small
\centering
\def\arraystretch{1.2}
\setlength\tabcolsep{7pt}
\caption{Different weight $\beta$ (\wrongbinop{}).}
\vspace{-2mm}
\label{table:contrastive-wrongbinop}
\resizebox{\columnwidth}{!}{\begin{tabular}{@{}llrlrlr@{}}
  \toprule
  \multirow{2.5}{*}{\shortstack[l]{Weight $\beta$ of\\Contrastive Loss}} & \multicolumn{2}{c}{\texttt{cls}} & \multicolumn{2}{c}{\texttt{cls-loc}} & \multicolumn{2}{c}{\hspace{2mm}\texttt{cls-loc-rep}} \\
  \cmidrule(lr){2-3} \cmidrule(lr){4-5} \cmidrule(l){6-7}
  & P & R & P & R & P & R \\
  \midrule
  0 & 47.96 & 43.18 & 47.06 & 42.36 & 46.15 & 41.55 \\
  0.25 & 47.92 & 44.60 & 47.05 & 43.79 & 46.17 & 42.97 \\
  0.5 & 47.62 & 44.81 & 46.75 & 43.99 & 46.10 & 43.38 \\
  1 & 48.51 & 46.44 & 47.66 & 45.62 & 46.60 & 44.60 \\
  2 & 51.54 & 44.20 & 50.36 & 43.18 & \textbf{49.64} & 42.57 \\
  $\star$ 4 & \textbf{52.30} & 43.99 & \textbf{51.09} & 42.97 & \textbf{49.64} & 41.75 \\
  8 & 48.35 & 41.75 & 47.41 & 40.94 & 46.70 & 40.33 \\
  16 & 50.23 & 43.79 & 49.30 & 42.97 & 48.36 & 42.16 \\
  \bottomrule
\end{tabular}}
      \vspace{1mm}
    \end{minipage}
    \medskip
    \begin{minipage}[t]{\columnwidth}
      \small
\centering
\def\arraystretch{1.2}
\setlength\tabcolsep{7pt}
\caption{Different weight $\beta$ (\argswap{}).}
\vspace{-2mm}
\label{table:contrastive-argswap}
\resizebox{\columnwidth}{!}{\begin{tabular}{@{}llrlrlr@{}}
  \toprule
  \multirow{2.5}{*}{\shortstack[l]{Weight $\beta$ of\\Contrastive Loss}} & \multicolumn{2}{c}{\texttt{cls}} & \multicolumn{2}{c}{\texttt{cls-loc}} & \multicolumn{2}{c}{\hspace{2mm}\texttt{cls-loc-rep}} \\
  \cmidrule(lr){2-3} \cmidrule(lr){4-5} \cmidrule(l){6-7}
  & P & R & P & R & P & R \\
  \midrule
  0 & 46.15 & 7.32 & 46.15 & 7.32 & 46.15 & 7.32 \\
  0.25 & 60.00 & 4.88 & 60.00 & 4.88 & 60.00 & 4.88 \\
  $\star$ 0.5 & 73.68 & 5.69 & 73.68 & 5.69 & 73.68 & 5.69 \\
  1 & 69.57 & 6.50 & 69.57 & 6.50 & 69.57 & 6.50 \\
  2 & 63.64 & 5.69 & 63.64 & 5.69 & 54.55 & 4.88 \\
  4 & 72.73 & 6.50 & 72.73 & 6.50 & 72.73 & 6.50 \\
  8 & 76.47 & 5.28 & 76.47 & 5.28 & 76.47 & 5.28 \\
  16 & \textbf{86.67} & 5.28 & \textbf{86.67} & 5.28 & \textbf{86.67} & 5.28 \\

  \bottomrule
\end{tabular}}
    \end{minipage}
  \end{minipage}\hfill
  \begin{minipage}[b]{0.48\columnwidth}
    \begin{minipage}[c]{\columnwidth}
      \small
\centering
\def\arraystretch{1.2}
\setlength\tabcolsep{7pt}
\caption{Different task order (\varmisuse{}).}
\vspace{-2mm}
\label{table:hierarchy-varmisuse}
\resizebox{\columnwidth}{!}{\begin{tabular}{@{}llrlrlr@{}}
  \toprule
  \multirow{3}{*}{Task Order} & \multicolumn{2}{c}{\texttt{cls}} & \multicolumn{2}{c}{\texttt{cls-loc}} & \multicolumn{2}{c}{\hspace{2mm}\texttt{cls-loc-rep}} \\
  \cmidrule(lr){2-3} \cmidrule(lr){4-5} \cmidrule(l){6-7}
  & P & R & P & R & P & R \\
  \midrule
  No Hierarchy & 51.82 & 16.96 & 50.91 & 16.67 & 48.18 & 15.77 \\
  $\star$ $cls$, $loc$, $rep$ & \textbf{64.79} & 13.69 & \textbf{61.97} & 13.10 & \textbf{56.34} & 11.90 \\
  $cls$, $rep$, $loc$ & 57.53 & 12.50 & 54.79 & 11.90 & 54.79 & 11.90 \\
  $loc$, $cls$, $rep$ & 53.66 & 13.10 & 50.00 & 12.20 & 47.56 & 11.61 \\
  $loc$, $rep$, $cls$ & 57.69 & 13.39 & 53.85 & 12.50 & 51.28 & 11.90 \\
  $rep$, $cls$, $loc$ & 51.69 & 13.69 & 49.44 & 13.10 & 47.19 & 12.50 \\
  $rep$, $loc$, $cls$ & 52.38 & 13.10 & 50.00 & 12.50 & 46.43 & 11.61 \\
  \bottomrule
\end{tabular}}
      \vspace{6.8mm}
    \end{minipage}
    \medskip
    \begin{minipage}[c]{\columnwidth}
      \small
\centering
\def\arraystretch{1.2}
\setlength\tabcolsep{7pt}
\caption{Different task order (\wrongbinop{}).}
\vspace{-2mm}
\label{table:hierarchy-wrongbinop}
\resizebox{\columnwidth}{!}{\begin{tabular}{@{}llrlrlr@{}}
  \toprule
  \multirow{3}{*}{Task Order} & \multicolumn{2}{c}{\texttt{cls}} & \multicolumn{2}{c}{\texttt{cls-loc}} & \multicolumn{2}{c}{\hspace{2mm}\texttt{cls-loc-rep}} \\
  \cmidrule(lr){2-3} \cmidrule(lr){4-5} \cmidrule(l){6-7}
  & P & R & P & R & P & R \\
  \midrule
  No Hierarchy & 48.70 & 45.62 & 47.83 & 44.81 & 46.52 & 43.58 \\
  $cls$, $loc$, $rep$ & 49.43 & 44.40 & 48.30 & 43.38 & 47.39 & 42.57 \\
  $cls$, $rep$, $loc$ & 48.29 & 43.18 & 47.38 & 42.36 & 46.01 & 41.14 \\
  $loc$, $cls$, $rep$ & 49.66 & 44.20 & 48.74 & 43.38 & 47.60 & 42.36 \\
  $loc$, $rep$, $cls$ & 46.44 & 46.44 & 45.42 & 45.42 & 44.81 & 44.81 \\
  $rep$, $cls$, $loc$ & 50.82 & 44.20 & 49.88 & 43.38 & 48.71 & 42.36 \\
  $\star$ $rep$, $loc$, $cls$ & \textbf{52.30} & 43.99 & \textbf{51.09} & 42.97 & \textbf{49.64} & 41.75 \\
  \bottomrule
\end{tabular}}
      \vspace{5.3mm}
    \end{minipage}
    \medskip
    \begin{minipage}[b]{\columnwidth}
      \small
\centering
\def\arraystretch{1.2}
\setlength\tabcolsep{7pt}
\caption{Different task order (\argswap{}).}
\vspace{-2mm}
\label{table:hierarchy-argswap}
\resizebox{\columnwidth}{!}{\begin{tabular}{@{}llrlrlr@{}}
  \toprule
  \multirow{3}{*}{Task Order} & \multicolumn{2}{c}{\texttt{cls}} & \multicolumn{2}{c}{\texttt{cls-loc}} & \multicolumn{2}{c}{\hspace{2mm}\texttt{cls-loc-rep}} \\
  \cmidrule(lr){2-3} \cmidrule(lr){4-5} \cmidrule(l){6-7}
  & P & R & P & R & P & R \\
  \midrule
  No Hierarchy & 61.29 & 7.72 & 61.29 & 7.72 & 61.29 & 7.72 \\
  $cls$, $loc$, $rep$ & \textbf{94.12} & 6.50 & \textbf{94.12} & 6.50 & \textbf{88.24} & 6.10 \\
  $cls$, $rep$, $loc$ & 53.57 & 6.10 & 53.57 & 6.10 & 53.57 & 6.10 \\
  $\star$ $loc$, $cls$, $rep$ & 73.68 & 5.69 & 73.68 & 5.69 & 73.68 & 5.69 \\
  $loc$, $rep$, $cls$ & 55.56 & 6.10 & 55.56 & 6.10 & 55.56 & 6.10 \\
  $rep$, $cls$, $loc$ & 76.47 & 5.28 & 76.47 & 5.28 & 76.47 & 5.28 \\
  $rep$, $loc$, $cls$ & 63.64 & 5.69 & 63.64 & 5.69 & 63.64 & 5.69 \\
  \bottomrule
\end{tabular}}
    \end{minipage}
  \end{minipage}
\end{table*}
\clearpage

\section{Case Studies of Inspected Warnings}
\label{sec:cases}
In the following we present case studies of the warnings we inspect in \cref{sec:scan}. We showcase representative bugs and code quality issues raised by our models. Further, we also provide examples of false positives and discuss potential causes for the failure. We visualise the bug location with \mycolorbox{mypurple} and the repair token with \mycolorbox{myyellow}.

\subsection{\varmisuse{}: bug in repository \mytexttt{aleju/imgaug}}
Our bug detector model correctly identifies the redundant check on \mytexttt{x\_px} instead of \mytexttt{y\_px}.
\begin{figure}[H]
\begin{lstlisting}[language=Python]
    def translate(self, x_px, (*@\hlrep{y\_px}@*)):
        if x_px < 1e-4 or x_px > 1e-4 or y_px < 1e-4 or (*@\hlloc{x\_px}@*) > 1e-4:
            matrix = np.array([[1, 0, x_px], [0, 1, y_px], [0, 0, 1]], dtype=np.float32)
            self._mul(matrix)
        return self
\end{lstlisting}
\end{figure}

\subsection{\varmisuse{}: bug in repository \mytexttt{babelsberg/babelsberg-r}}
The model identifies that \mytexttt{w\_read} was already checked but not \mytexttt{w\_write}.
\begin{figure}[H]
\begin{lstlisting}[language=Python]
    def test_pipe(self, space):
        w_res = space.execute("""
        return IO.pipe
        """)
        w_read, (*@\hlrep{w\_write}@*) = space.listview(w_res)
        assert isinstance(w_read, W_IOObject)
        assert isinstance((*@\hlloc{w\_read}@*), W_IOObject)
        w_res = space.execute("""
        r, w, r_c, w_c = IO.pipe do |r, w|
          r.close
          [r, w, r.closed?, w.closed?]
        end
        return r.closed?, w.closed?, r_c, w_c
        """)
        assert self.unwrap(space, w_res) == [True, True, True, False]
\end{lstlisting}
\end{figure}

\subsection{\varmisuse{}: code quality issue in repository \mytexttt{JonnyWong16/plexpy}}
The model proposes to replace \mytexttt{c} by \mytexttt{snowman}, since \mytexttt{snowman} is otherwise unused. Even though this replacement does not suggest a bug, the warning remains useful as the unused variable \mytexttt{snowman} must be considered a code quality issue.
\begin{figure}[H]
\begin{lstlisting}[language=Python, upquote=true]
    def test_ensure_ascii_still_works(self):
        # in the ascii range, ensure that everything is the same
        for c in map(unichr, range(0, 127)):
            self.assertEqual(
                json.dumps(c, ensure_ascii=False),
                json.dumps(c))
        (*@\hlrep{snowman} @*)= u(*@\texttt{'}@*)\N{SNOWMAN}(*@\texttt{'} @*)
        self.assertEqual(
            json.dumps(c, ensure_ascii=False), 
            (*@ \texttt{'"'} @*) + (*@\hlloc{c}@*) + (*@ \texttt{'"'}) @*)
\end{lstlisting}
\end{figure}

\subsection{\varmisuse{}: false positive in repository \mytexttt{ceache/treadmill}}
The model proposes to replace \mytexttt{fmt} with \mytexttt{server}, although the surrounding code clearly implies that \mytexttt{server} is \mytexttt{None} at this point in the program. Therefore, we consider it as a false positive. In this case, the two preceding method calls with \mytexttt{\_server} in their name, were given the \mytexttt{server} variable as second argument. This may have affected the prediction of the model, disregarding the surrounding conditional statement with respect to \mytexttt{server}.
\begin{figure}[H]
\begin{lstlisting}[language=Python]
    def server_cmd((*@\hlrep{\texttt{server}}@*), reason, fmt, clear):
        """Manage server blackout."""
        if server is not None:
            if clear:
                _clear_server_blackout(context.GLOBAL.zk.conn, server)
            else:
                _blackout_server(context.GLOBAL.zk.conn, server, reason)
        else:
            _list_server_blackouts(context.GLOBAL.zk.conn, (*@\hlloc{\texttt{fmt}}@*))
\end{lstlisting}
\end{figure}

\subsection{\wrongbinop{}: bug in repository \mytexttt{Amechi101/concepteur-market-app}}
The model detects the presence of the string formatting literal \mytexttt{\%s} in the string and in consequence raises a warning about a wrong binary operator.
\begin{figure}[H]
\begin{lstlisting}[language=Python]
    +  *  -  /  (*@\hlrep{\texttt{\%}}@*)
    def buildTransform(inputProfile, outputProfile, inMode, outMode, 
        renderingIntent=INTENT_PERCEPTUAL, flags=0):

        if not isinstance(renderingIntent, int) or not (0 <= renderingIntent <=3):
            raise PyCMSError((*@\texttt{"}@*)renderingIntent must be an integer between 0 and 3(*@\texttt{"}@*))

        if not isinstance(flags, int) or not (0 <= flags <= _MAX_FLAG):
            raise PyCMSError((*@\texttt{"}@*)flags must be an integer between 0 and %s(*@\texttt{"}@*) (*@\hlloc{\texttt{+}}@*) _MAX_FLAG)

        try:
            if not isinstance(inputProfile, ImageCmsProfile):
                inputProfile = ImageCmsProfile(inputProfile)
            if not isinstance(outputProfile, ImageCmsProfile):
                outputProfile = ImageCmsProfile(outputProfile)
            return ImageCmsTransform(inputProfile, outputProfile, inMode, outMode,
                                     renderingIntent, flags=flags)
        except (IOError, TypeError, ValueError) as v:
            raise PyCMSError(v)
\end{lstlisting}
\end{figure}

\subsection{\wrongbinop{}: bug in repository \mytexttt{maestro-hybrid-cloud/heat}}
The model correctly raises a warning since the comparison must be a containment check instead of an equality check.
\begin{figure}[H]
\begin{lstlisting}[language=Python]
    ==  !=  (*@\texttt{is}@*)  (*@\texttt{is not}@*)  <  <=  (*@\texttt{>}@*)  >=  (*@\hlrep{\texttt{in}}@*)  (*@\texttt{not in}@*)
    def suspend(self):
        # No need to suspend if the stack has been suspended
        if self.state (*@\hlloc{\texttt{==}}@*) (self.SUSPEND, self.COMPLETE):
            LOG.info(_LI((*@\texttt{'}@*)%s is already suspended(*@\texttt{'}@*)), six.text_type(self))
            return

        self.updated_time = datetime.datetime.utcnow()
        sus_task = scheduler.TaskRunner(
            self.stack_task,
            action=self.SUSPEND,
            reverse=True,
            error_wait_time=cfg.CONF.error_wait_time)
        sus_task(timeout=self.timeout_secs())
\end{lstlisting}
\end{figure}

\subsection{\wrongbinop{}: code quality issue in repository \mytexttt{tomspur/shedskin}}
The model identifies the unconventional use of the \mytexttt{!=} operator when comparing with \mytexttt{None}.
\begin{figure}[H]
\begin{lstlisting}[language=Python]
    ==  !=  (*@\texttt{is}@*)  (*@\hlrep{\texttt{is not}}@*)  <  <=  (*@\hlrep{\texttt{>}}@*)  >=  (*@\texttt{in}@*)  (*@\texttt{not in}@*)
    def getFromEnviron():
        if HttpProxy.instance is not None:
            return HttpProxy.instance
        url = None
        for key in ((*@\texttt{'}@*)http_proxy(*@\texttt{'}@*), (*@\texttt{'}@*)https_proxy(*@\texttt{'}@*)):
            url = os.environ.get(key)
            if url: break
        if not url:
            return None
        dat = urlparse(url)
        port = 80 if dat.scheme == (*@\texttt{'}@*)http(*@\texttt{'}@*) else 443
        if dat.port (*@\hlloc{\texttt{!=}}@*) None: port = int(dat.port)
        host = dat.hostname
        return HttpProxy((host, port), dat.username, dat.password)
\end{lstlisting}
\end{figure}

\subsection{\wrongbinop{}: false positive in repository \mytexttt{wechatpy/wechatpy}}
Our model mistakenly raises a \wrongbinop{} warning with the \mytexttt{==} operator and proposes to replace it with \mytexttt{>} operator. In this case, the log message below the conditional check may have triggered the warning.
\begin{figure}[H]
\begin{lstlisting}[language=Python]
    ==  !=  (*@\texttt{is}@*)  (*@\texttt{is not}@*)  <  <=  (*@\hlrep{\texttt{>}}@*)  >=  (*@\texttt{in}@*)  (*@\texttt{not in}@*)
    def add_article(self, article):
        if len(self.articles) (*@\hlloc{\texttt{==}}@*) 10:
            raise AttributeError((*@\texttt{"}@*)Can(*@\texttt{'}@*)t add more than 10 articles in an ArticlesReply(*@\texttt{"}@*))
        articles = self.articles
        articles.append(article)
        self.articles = articles
\end{lstlisting}
\end{figure}

\subsection{\argswap{}: bug in repository \mytexttt{sgiavasis/nipype}}
Our model identifies the invalid use of the NumPy function \mytexttt{np.savetxt(streamlines, out\_file + '.txt')}\footnote{NumPy Documentation, \url{https://numpy.org/doc/stable/reference/generated/numpy.savetxt.html}}, which expects first the file name, \ie, \mytexttt{out\_file + '.txt'} in this case.
\begin{figure}[H]
\begin{lstlisting}[language=Python]
    def _trk_to_coords(self, in_file, out_file=None):
        from nibabel.trackvis import TrackvisFile
        trkfile = TrackvisFile.from_file(in_file)
        streamlines = trkfile.streamlines

        if out_file is None:
            out_file, _ = op.splitext(in_file)

        np.savetxt((*@\hlloc{streamlines}@*), (*@\hlrep{out\_file}@*) + '.txt')
        return out_file + '.txt'
\end{lstlisting}
\end{figure}

\subsection{\argswap{}: false positive in repository \mytexttt{davehunt/bedrock}}
Our model mistakenly raises an argument swap warning with the \mytexttt{SpacesPage} constructor. In fact, with this specific repository our model repeatedly raised issues at similar code locations where the Selenium library is used. This is likely due to not having encountered similar code during training and hence due to lack of repository-specific information.
\begin{figure}[H]
\begin{lstlisting}[language=Python]
    @pytest.mark.nondestructive
    def test_spaces_list(base_url, selenium):
        page = SpacesPage((*@\hlloc{base\_url}@*), (*@\hlrep{selenium}@*)).open()
        assert page.displayed_map_pins == len(page.spaces)
        for space in page.spaces:
            space.click()
            assert space.is_selected
            assert space.is_displayed
            assert 1 == page.displayed_map_pins
\end{lstlisting}
\end{figure}

\section{Bug Reports to the Developers}
\label{sec:reports}
We report a number of bugs found during our manual inspection as pull requests to the developers. For forked repositories, we trace the buggy code in the original repository. If the original repository has the same code, we create a bug report in the original repository. Otherwise, we do not report the bug. We also found that 7 bugs are already fixed in the latest version of the repository. The links to the pull requests are listed below. We also mark the pull requests for which we received a confirmation from the developers before the deadline for the final version of this paper (two days after we reported them).

\varmisuse{}: \\
(merged) \url{https://github.com/numpy/numpy/pull/21764} \\
(merged) \url{https://github.com/frappe/erpnext/pull/31372} \\
(merged) \url{https://github.com/spirali/kaira/pull/31} \\
(merged) \url{https://github.com/pyro-ppl/pyro/pull/3107} \\
(merged) \url{https://github.com/nest/nestml/pull/789} \\
(merged) \url{https://github.com/cupy/cupy/pull/6786} \\
(merged) \url{https://github.com/funkring/fdoo/pull/14} \\
(confirmed) \url{https://github.com/apache/airflow/pull/24472} \\
\url{https://github.com/topazproject/topaz/pull/875} \\
\url{https://github.com/inspirehep/inspire-next/pull/4188} \\
\url{https://github.com/CloCkWeRX/rabbitvcs-svn-mirror/pull/6} \\
\url{https://github.com/amonapp/amon/pull/219} \\
\url{https://github.com/mjirik/io3d/pull/9} \\
\url{https://github.com/jhogsett/linkit/pull/30} \\
\url{https://github.com/aleju/imgaug/pull/821} \\
\url{https://github.com/python-diamond/Diamond/pull/765} \\
\url{https://github.com/python/cpython/pull/93935} \\
\url{https://github.com/orangeduck/PyAutoC/pull/3} \\
\url{https://github.com/damonkohler/sl4a/pull/332} \\
\url{https://github.com/vyrus/wubi/pull/1} \\
\url{https://github.com/shon/httpagentparser/pull/89} \\
\url{https://github.com/midgetspy/Sick-Beard/pull/991} \\
\url{https://github.com/sgala/gajim/pull/3} \\
\url{https://github.com/tensorflow/tensorflow/pull/56468} \\
\\
\wrongbinop{}: \\
(merged) \url{https://github.com/python-pillow/Pillow/pull/6370} \\
(merged) \url{https://github.com/funkring/fdoo/pull/15} \\
(false positive) \url{https://github.com/kovidgoyal/calibre/pull/1658} \\
\url{https://github.com/kbase/assembly/pull/327} \\
\url{https://github.com/maestro-hybrid-cloud/heat/pull/1} \\
\url{https://github.com/gramps-project/gramps/pull/1380} \\
\url{https://github.com/scikit-learn/scikit-learn/pull/23635} \\
\url{https://github.com/pupeng/hone/pull/1} \\
\url{https://github.com/edisonlz/fruit/pull/1} \\
\url{https://github.com/certsocietegenerale/FIR/pull/275} \\
\url{https://github.com/MediaBrowser/MediaBrowser.Kodi/pull/117} \\
\url{https://github.com/sgala/gajim/pull/4} \\
\url{https://github.com/mapsme/omim/pull/14185} \\
\url{https://github.com/tensorflow/tensorflow/pull/56471} \\
\url{https://github.com/catapult-project/catapult-csm/pull/2} \\
\\
\argswap{}: \\
(merged) \url{https://github.com/clinton-hall/nzbToMedia/pull/1889} \\
(merged) \url{https://github.com/IronLanguages/ironpython3/pull/1495} \\
(false positive) \url{https://github.com/python/cpython/pull/93869} \\
\url{https://github.com/google/digitalbuildings/pull/646} \\
\url{https://github.com/quodlibet/mutagen/pull/563} \\
\url{https://github.com/nipy/nipype/pull/3485} \\

\end{document}